\documentclass[lettersize,journal]{IEEEtran}
\usepackage{amsmath,amsfonts}
\usepackage{algorithmic}
\usepackage{algorithm}
\usepackage{array}
\usepackage[caption=false,font=normalsize,labelfont=sf,textfont=sf]{subfig}
\usepackage{textcomp}
\usepackage{stfloats}
\usepackage{url}
\usepackage{verbatim}
\usepackage{graphicx}
\usepackage{cite}

\usepackage{multirow}
\usepackage{mathtools}
\usepackage{booktabs}
\usepackage{makecell}

\begin{document}

\title{Guided Image-to-Image Translation by Discriminator-Generator Communication}

\author{Yuanjiang Cao, Lina Yao, \IEEEmembership{Senior Member,~IEEE}, Le Pan, Quan Z. Sheng, \IEEEmembership{Member,~IEEE}, and Xiaojun Chang, \IEEEmembership{Senior Member,~IEEE} 

\thanks{Yuanjiang Cao, Lina Yao, and Le Pan are with the School of Computer Science and Engineering, University of New South Wales, Sydney, NSW, AUS 2052. Email: yuanjiang.cao@unsw.edu.au, lina.yao@unsw.edu.au, le.pan@unsw.edu.au}
\thanks{Lina Yao is with CSIRO's Data61. Email: lina.yao@data61.csiro.au}
\thanks{Quan Z. Sheng (Michael Sheng) is with Department of Computing, Macquarie University, Sydney, NSW, AUS. Email: michael.sheng@mq.edu.au}
\thanks{Xiaojun Chang is with Faculty of Engineering and Information Technology, University of Technology Sydney, Sydney, NSW, AUS. Email:XiaoJun.Chang@uts.edu.au}
}

\markboth{Journal of \LaTeX\ Class Files,~Vol.~14, No.~8, August~2021}%
{Shell \MakeLowercase{\textit{et al.}}: A Sample Article Using IEEEtran.cls for IEEE Journals}


\maketitle

\begin{abstract}
The goal of Image-to-image (I2I) translation is to transfer an image from a source domain to a target domain, which has recently drawn increasing attention. One major branch of this research is to formulate I2I translation based on Generative Adversarial Network (GAN). As a zero-sum game, GAN can be reformulated as a Partially-observed Markov Decision Process (POMDP) for generators, where generators cannot access full state information of their environments. This formulation illustrates the information insufficiency in the GAN training. To mitigate this problem, we propose to add a communication channel between discriminators and generators. 
We explore multiple architecture designs to integrate the communication mechanism into the I2I translation framework. To validate the performance of the proposed approach, we have conducted extensive experiments on various benchmark datasets. The experimental results confirm the superiority of our proposed method.

\end{abstract}

\section{Introduction}
Image-to-image (I2I) translation aims to transform an image from one domain to another \cite{zheng2021one,zheng2022asynchronous,chen2019quality,pang2021image,huang2021multi}. It has been broadly used in applications such as animation translation \cite{kimUgatitUnsupervisedGenerative2019}, image super-resolution\cite{dong2015image}, style transfer\cite{gatysImageStyleTransfer2016a}, image segmentation\cite{guo2020gan}, sketch-to-image translation \cite{huang2021multi}, and Sim2Real for robotics\cite{rao2020rl}. 
I2I translation model selection depends on whether data are paired. Early research on paired I2I translation can be modelled through supervised learning. For instance, generative models (VAE or auto-regressive models, etc.) are naturally applied on I2I translation tasks with paired target images, such as converting a sketch into a commodity image\cite{huangMultimodalUnsupervisedImagetoimage2018}. While obtaining paired images can be difficult and expensive (such as segmentation) or even ill-defined (such as substituting a cat with a dog), research of unpaired (unsupervised) I2I translation thrives to bypass this difficulty. It captures the supervision signal from a domain distribution via a neural model\cite{zhuUnpairedImagetoimageTranslation2017}. Combined with the effective Generative Adversarial Network (GAN) \cite{goodfellow2020generative} where discriminators and generators are created and converge to a Nash-equilibrium of the zero-sum game, more powerful models for unpaired I2I translation have been developed. 



\begin{figure}[t]
\centering
  \includegraphics[width=0.48\textwidth]{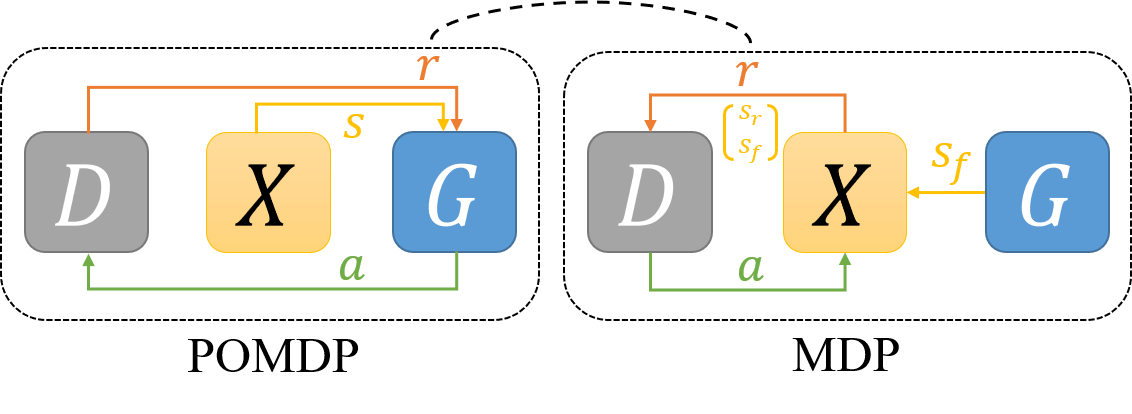}
  \label{fig:motivation-pomdp}
\caption{The defined Partially-observed Markov Decision Process (POMDP) based on GAN. $D, G, X$ represent a discriminator, a generator and a batch of images from a dataset, respectively. $s$ is state, $s_r$ is an image from the real domain, and $s_f$ is an image from the fake domain. $r$ is the reward for an agent, and $a$ is the action for the agent. The black dashed arc in this figure shows the comparison of the state source between the MDP of the discriminator and the POMDP of the generator.}
\end{figure}

Former studies in I2I mainly focus on constraining the parameters of the GAN through consistencies. However, they neglect \textbf{information insufficiency} between discriminators and generators. We demonstrate this disadvantage of the training process of the GAN from a decision-making perspective, as shown in Figure \ref{fig:motivation-pomdp}. The game in the GAN is formed between a discriminator and a generator, whose main objective is the adversarial loss.
Sub-modules of the GAN can be naturally regarded as agents in this game. For each agent, other sub-modules can be viewed as environments, so we can define MDPs based on this.
In Figure \ref{fig:motivation-pomdp}, we define two MDPs by taking the discriminator and the generator as agents separately. The generator's MDP is different from the discriminator's because the generator cannot observe the full state of its environment. A partial state leads to information insufficiency for the generator, and this difference makes the generator's MDP a Partially-observed Markov Decision Process (POMDP).
To address this problem, we convert the POMDP to an MDP by introducing a state from the discriminator to the generator during image translation. We treat this state as a communication scheme inspired by the communication of multi-agent RL.


From the perspective of the communication scheme, two conventional models, the observation-respond scheme and the coordination scheme 
are mostly used in previous research\cite{gronauerMultiagentDeepReinforcement2022a}. The former scheme treats communication as observations, where other agents' behaviour data are coded in the state, and the agent reasons the behaviour through an internal model. The latter explicitly models the communication as discrete or continuous vectors passed among the agents where the vectors are learned. As the observation-respond scheme does not change the setting of POMDP for the generator, our study uses the coordination scheme where the discriminator propagates a continuous message vector to the generator.

A message vector can be viewed as a guidance vector as well, and the guidance for the generator in I2I translation has been explored by previous studies \cite{mejjati2018unsupervised, emami2020spa}. In \cite{mejjati2018unsupervised}, an attention mask is extracted from a third network apart from generators and discriminators, and the inner product of the image and the mask are taken as input for the training of both discriminators and generators. 
The mask generator is trained for a certain epoch to extract the attention to stabilize the training process. 
Instead of computing attention mask from an extra neural network, researchers \cite{emami2020spa} leverage the aggregation of intermediate feature maps of the discriminators directly.
The masks are integrated into the input of the CycleGAN model with pixel-level constraints on intermediate feature maps.
These models apply the guidance on the input space which might lead to instability of training, where extra constraints like training the model for certain epochs are required. Our model learns the message vector that can adaptively control the amount of information without these specific constraints.


Considering all the above perspectives, we devise a novel guided Image-to-Image translation model.
This model addresses the disadvantage of the POMDP formulation by creating communication channels.
In our model, the channel is defined as a continuous communication vector sent by the discriminator to the generator. The communication vector is estimated and learned with normal GAN loss.
We evaluate our model on classic I2I datasets. The experimental results show that learned communication can aid higher-quality translation in multiple configurations. In summary, our contributions are three-fold:
\begin{itemize}
\item
To the best of our knowledge, we are the first to view GAN as a POMDP for generators. Based on
this premise, we further explore the communication setting for unpaired I2I translation.

\item
 We design the communication scheme in multiple classic GAN models and investigate various communication schemes for unpaired I2I translation.

\item
We conduct experiments on classic datasets of this research area. Experiments show that our model can enhance the performance of the current I2I translation model with slight changes in the model architecture.

\end{itemize}

\section{Related Work}
\textbf{Image-to-Image translation} attracts hot attention in recent years, applied to a wide range of tasks, such as style transfer\cite{isolaImagetoimageTranslationConditional2017}, super-resolution synthesis\cite{chenAttentionganObjectTransfiguration2018}, and domain adaptation\cite{zhuUnpairedImagetoimageTranslation2017}. Common network backbones are conditional generation models, conditional GAN-based models \cite{isolaImagetoimageTranslationConditional2017,wangHighresolutionImageSynthesis2018} or non-GAN based model \cite{mechrezContextualLossImage2018}. GAN-based models can be further separated by one-sided model~\cite{benaimOnesidedUnsupervisedDomain2017} and two-sided model~\cite{zhuUnpairedImagetoimageTranslation2017,yiDualganUnsupervisedDual2017,royerXganUnsupervisedImagetoimage2020,choiStarganV2Diverse2020,liuUnsupervisedImagetoimageTranslation2017,leeDiverseImagetoimageTranslation2018}. 
One-sided models focus on learning without cycle-consistency, while two-sided models leverage cycle-consistency as a strong regularization for generators. 
Based on Cycle-consistency, two sided models conduct research on disentangle representation \cite{huangMultimodalUnsupervisedImagetoimage2018,leeDiverseImagetoimageTranslation2018}, diverse generation \cite{choiStarganUnifiedGenerative2018, choiStarganV2Diverse2020, royerXganUnsupervisedImagetoimage2020} and auxilary agumentation \cite{jiangSaliencyGuidedImageTranslation2021, zhang2020cross, chenAttentionganObjectTransfiguration2018}.
In one-sided models, Benaim and Wolf~\cite{benaimOnesidedUnsupervisedDomain2017} capture the distance similarities in source and target domains. CUT~\cite{parkContrastiveLearningUnpaired2020} introduces feature contrast, which achieves good performance. Other extensions include new losses inspired by sample-sample relationship~\cite{mechrezContextualLossImage2018}, visual feature augumentation losses~\cite{johnsonPerceptualLossesRealtime2016}, novel normalization~\cite{kimUgatitUnsupervisedGenerative2019,parkSemanticImageSynthesis2019b}. These works explore representation learning without changing the GAN formulation.

The guidance for the generator in I2I translation has been explored by previous studies \cite{mejjati2018unsupervised,emami2020spa}. In \cite{mejjati2018unsupervised}, the guidance is in the form of an attention mask. The masks extracted from an extra network are trained in tandem with generators. The masks also provide guidance for discriminators.
Extra hand-designed rules exist in the training. The mask generators are trained for 30 epochs before feeding into discriminators to stablize the training process. 
The guidance in \cite{emami2020spa} is intermediate feature maps of discriminators. These features are processed by hand-designed aggregation flow and then fed into generators directly. 
In contrast, our model learns the guidance in the form of continuous communication vectors without gradient freezing, where the discriminator can learn to propagate useful information to generators and leave out the noisy features. Our learned vectors do not require a predefined training scheme. Also, our model can scale to different architectures easily.


\noindent\textbf{Communication in Multi-agent Reinforcement Learning} Communication among multi-agents is one of the core challenges of machine intelligence \cite{gronauerMultiagentDeepReinforcement2022a}. One common framework for multi-agent communication in reinforcement learning has been proposed in \cite{oliehoek2016concise}, where communication is discussed under the decentralized POMDP. In this setting, each agent can only receive a partial state of the environment. It defines critical concepts like implicit and explicit communication, the delayed communication and the cost of communication. 
\cite{foerster2016learning} proposes discrete communication in MARL. In contrast, \cite{sukhbaatar2016learning} utilizes continuous communication protocol, where communication vectors are merged by computing the average of the vectors. In \cite{pengMultiagentBidirectionallyCoordinatedNets2017a}, a bidirectional-coordinated network is introduced where heterogeneous communication can be handled by a sequence model with the Backpropaget through time (BPTT) scheme. This model is applied to StarCraft AI.

Apart from the message type, the strategy of target selection is also important in the communication research \cite{gronauerMultiagentDeepReinforcement2022a}. Attention mechanism is introduced in order to select which agent to communicate with. In \cite{jiang2018learning}, the model ATOC applies an attention model on the sender to communicate dynamically. \cite{hoshenVAINAttentionalMultiagent2017a} proposes a vertex attention interaction network (VAIN) to focus only on relevant agents. Agents in the TarMAC \cite{dasTarMACTargetedMultiAgent2019a} generate messages as well as signatures to actively choose which agent to communicate with. The signature is a vector that acts as a key or query that will be matched with the receiver's key by a distance score. IC3Net \cite{singh2018learning} pipes messages into a gate to allow an agent to block certain communication.

\section{Methodology}
\label{sec:methodology}


In this section, we first demonstrate the formulation of GAN as a POMDP for generators. Then we introduce the design of the proposed communication module.

\subsection{GAN as POMDP for generators}
\label{sec:methodology-pomdp}

We demonstrate the infromation insufficiency in Figure \ref{fig:motivation-pomdp}. We view the training process of GAN as a POMDP for the generator.
There are three submodules in the training process of the GAN, the discriminator, the generator and  the dataset. 
We create two MDPs for different agents, the generator amd the discriminator.
When we take the discriminator as the agent, states and rewards come from the dataset.
The fake image generated from the generator forms a part of the image dataset, which means that given a fake image, the discriminator and the generator are independent of each other. 
The independence determines that the discriminator gets full observation of its environment. 
In contrast, the state and reward for a generator come from the dataset and the discriminator separately when we define the MDP for generators.
The training signal for generators comes from discriminators, which must be defined as rewards for generators. 
The state observed by the generator are purely real images from the dataset, containing nothing about the state of the discriminator. 
Moreover, the parameters of the discriminator change over training steps. The ever-changing parameters form a distribution of the environment. 
Above setting satisfies the definition of a POMDP \cite{oliehoek2016concise}.

POMDP is a variant of MDP, which contains a set of variables representing the signal for learning and the interaction between agents and an environment. Specifically, it is formulated as a set $(S, A, T, R, \Omega, O, \gamma)$, and we show the items formulation list as follows:
\begin{itemize}
    \item $S$ is a set of states $s$ of the environment
    \item $A$ is a set of actions $a$ of the agents
    \item $T$ is the transition mechanism, which can be represented as a conditional probability of the next state given the current state and action $p(s^{\prime}|s, a)$
    \item $R: S \times A \rightarrow \mathcal{R}$ is the reward function
    \item $\Omega$ is a set of observations that can be the partition of states
    \item $O$ is the belief of the agent that represents the observation probabilities $p(o |s^{\prime}, a)$
    \item $\gamma$ is the discount factor of the reward
\end{itemize}

In the formulation of GAN, $S$ is the Cartesian product of the set of real images and the set of parameters of discriminator $D$, and $A$ is the set of generated images. Each episode of the one-step POMDP only contains one state $s$, and $s$ is not changed after the action. Therefore, in $T$, $s^{\prime} = s$. $R$ is formulated as the gradient propagate from the discriminator $R = \frac{\partial L_{gen}}{\partial D}$, $L_{gen}$ is the loss function for generators. $\Omega$ is the set of images observed by the generator, and $O$ is the probability $p(x|x,D,a)$.
For a discriminator, the game forms an one-step MDP. $S$ is the set of image-label pairs, and $A$ is the set of probability output, and $R$ is the gradient of adversarial loss.

This formulation is a detailed illustration of the communication between the discriminator to the generator in Figure \ref{fig:motivation-pomdp}.  

\subsection{Background of CycleGAN}

The I2I translation aims to learn one or more generators to translate images between one or multiple domains. We take the translation between two domains as an example. Suppose we have two domains $A, B \subset R^{C \times H \times W}$, where $C$ stands for the channel of image space, $H, W$ the height and the width. 
We hypothesize that all images in one domain satisfy one distribution defined on the image space. The distribution on domains is defined as $p_A$ and $p_B$, where image group $(x_A, x_B)$ are sampled. We define $x$ as an image sample without specification of domains. $p(\cdot)$ is the symbol of a certain probability distribution. Our model can be integrated into any GAN-based I2I translation model. In this study, we use CycleGAN~\cite{zhuUnpairedImagetoimageTranslation2017} as an example. First, we introduce the architecture of CycleGAN. It has two generators (i.e. $G_{AB}$ and $G_{BA}$) and two discriminators (i.e. $D_{A}$ and $D_{B}$). Generators take unpaired images $x_A \in A$ and $x_B \in B$ as input and generate samples in the other domain $B$ and $A$ correspondingly.

The objective of discriminators follows the traditional GAN routine~\cite{zhuUnpairedImagetoimageTranslation2017}, i.e., to classify whether a sample is a real or a generated image. This approach, combined with generators' adversarial loss, forms an adversarial min-max game. A discriminator and a generator form a pair in training, where they play the zero-sum min-max game. 
\begin{equation}
\label{eq:gan-loss}
\begin{aligned}
    \underset{G_A}{\operatorname{min}} \; \underset{D_A}{\operatorname{max}} \; L_{gan} \\
    L_{gan}(D_A, G_{BA}) = \mathop{\mathbb{E}}_{x_A \sim p_A} [-\operatorname{log}(D_A(x_A))] \\
                   + \mathop{\mathbb{E}}_{x_B \sim p_B} [-\operatorname{log}(1 - D_A(G_{BA}(x_B)))]
\end{aligned}
\end{equation}

Generators are trained with multiple objectives, cycle consistency, adversarial objective, and regularization objective. Following CycleGAN, we use the cycle consistency loss to keep the generators' output images with content consistency. The cycle consistency is a reconstruction $L_1$ loss by double mapping of input images. Take $x_A$ as an example, double mapping is to generate a sample $G_{AB} \in p_B$, and then map the sample back to the original domain $G_{BA}(G_{AB}) \in p_A$. Because the neural network is trained jointly in two directions, the cycle consistency allows the model to change the texture of the image while keeping the shape and geometry static. Although this constraint is strict and likely to lead to failure in large domain gaps~\cite{zhengSpatiallyCorrelativeLossVarious2021}, the constraint works well in our experiments.
\begin{equation}
\begin{aligned}
    L_{cyc} = \mathop{\mathbb{E}}_{x_A \sim p_A} ||x_A - G_{BA}(G_{AB}(x_A))||_1 \\
    + \mathop{\mathbb{E}}_{x_B \sim p_B} ||x_B - G_{AB}(G_{BA}(x_B))||_1
\end{aligned}
\end{equation}

\begin{figure}[t]
\centering
\subfloat[\label{fig:cycle-gan}]{
  \includegraphics[width=0.48\textwidth]{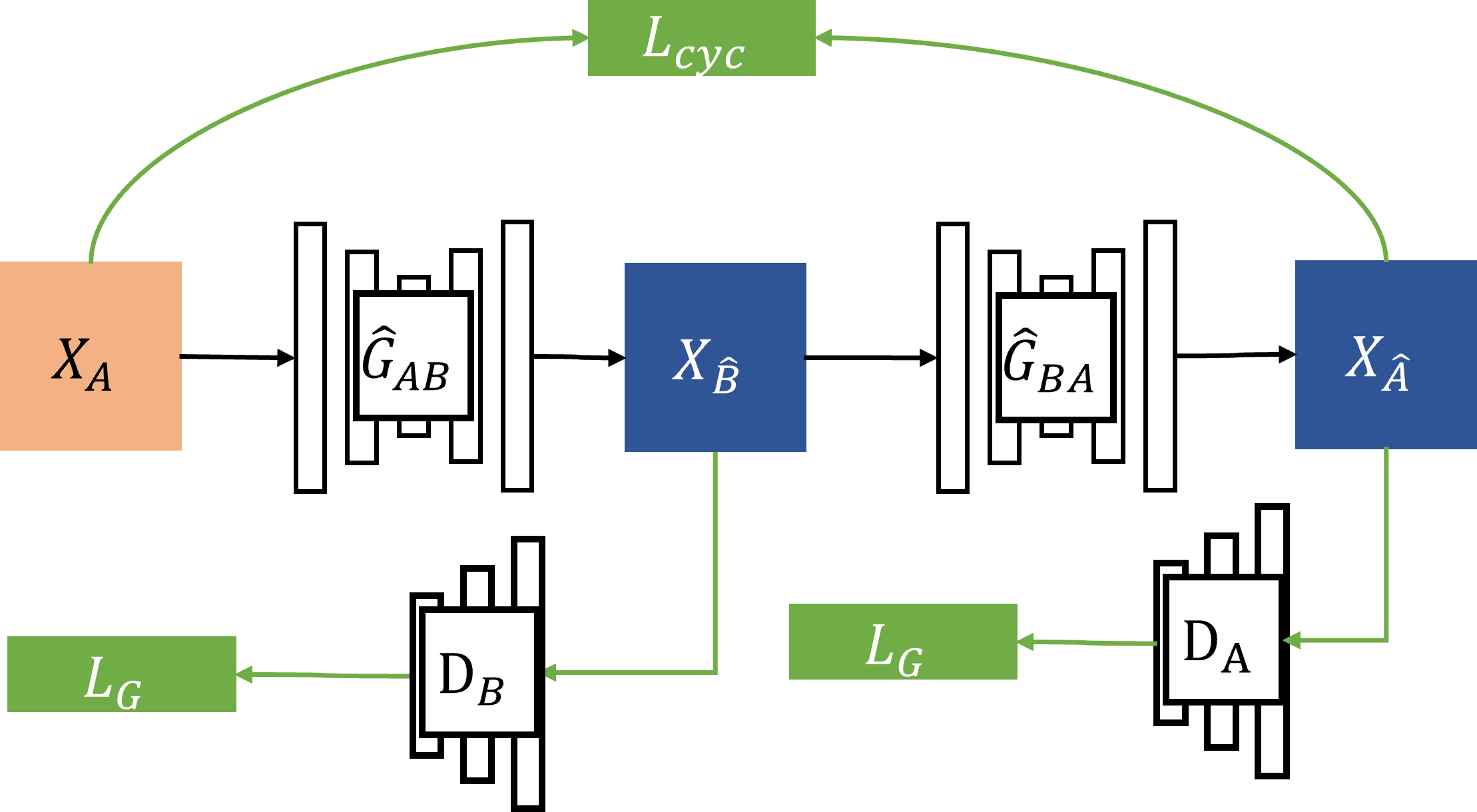}
 }

\hfill

\subfloat[\label{fig:guidance}]{
  \includegraphics[width=0.48\textwidth]{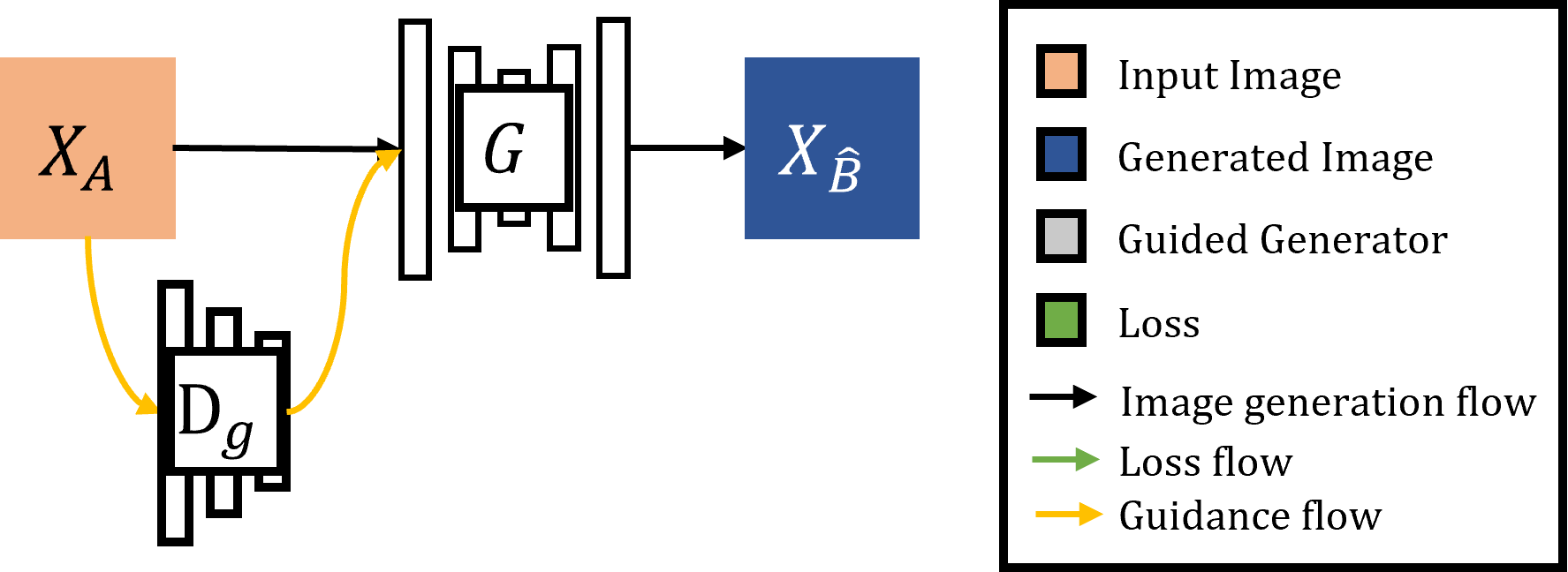}
}
\caption{The proposed model architecture. (a) shows the data flow of cycle consistent loss and adversarial loss, (b) shows how the communication module is integrated into the model.}
\end{figure}


 

\subsection{Guidance module in the GAN}

In this section, we develop our guided GAN model by modifying the formulation of the generator computation. Here we introduce a few new variables. The guidance representation is defined as $g$, which is a continuous vector. The model that outputs the guidance vector is defined as $H$. In contrast, in multi-agent RL, the communication graph is learned, where multiple iterations of communication are executed. The message or the communication is computed and propagated into each agent in every iteration. 
One key difference between the GAN setting and the multi-agent setting is the asymmetry of agents defined in the game. 
As shown in Fig \ref{fig:motivation-pomdp} and Section \ref{sec:methodology-pomdp}, the game can be viewed as an MDP for the discriminator. This means it receives the full state required to model the binary image classification (in CycleGAN, it is patch-based binary classification). Given the generated image, the objective of discriminators is independent of the generators, making it redundant to add new communication ways from generators to discriminators. The POMDP only provides implicit communication via reward for generators. Therefore, we only consider the communication for generators.

Given that the guidance is from discriminator to generator, the guidance model $H$ is overwritten as $D_g$. The guidance vector computed from the $D_g$ is sent to a generator. The guidance is computed in the following,

\begin{equation}
\begin{aligned}
    g = D_g(x) \\
    \hat{x} = G(x, g)
\end{aligned}
\end{equation}

Fig \ref{fig:cycle-gan} shows the architecutre of CycleGAN and Fig \ref{fig:guidance} illustrates how the output of $D_g$ is integrated into a CycleGAN model.

\subsection{Guidance model and Merge strategy}
The guidance model can be constructed in various ways.  We can define it as an output of one discriminator, multiple discriminators or combining the knowledge from trained models. For example, it can be the set of the discriminators of two domains as well as auxiliary classifiers in \cite{kimUgatitUnsupervisedGenerative2019}. In this subsection, we  first discuss the architecture of the guidance source, and then we introduce our design of how to merge the guidance vector into the generator.
We explore the guidance under two conditions: guidance from a single discriminator and guidance from multiple discriminators. We don't explore the auxiliary guidance because it requires extra parameters. 
We discuss the advantages and disadvantages of discriminator selection in the single discriminator scenario. In the I2I translation task, there is one discriminator for each domain. We refer to the two discriminators as the input domain discriminator and the output domain discriminator. The input and output are defined in respect of a certain generator. To address the POMDP issue, we need to use the discriminator that forms the game with the generator, the output domain discriminator, where communication vectors can be used as probes for the discriminator. While from the view of guidance, as mentioned in the literature \cite{mejjati2018unsupervised, emami2020spa}, the input domain discriminator extracts salient features after being trained on real and fake images in the input domain, which can provide useful information as well. We experiment with these two designs in the analysis in Section \ref{sec:exp}.

Different from single domain guidance, the other option is to merge the guidance from multiple discriminators. We discuss two approaches to address this issue. One is to directly average the guidance vectors \cite{sukhbaatar2016learning}. One variation of this is to learn a weighted average. The formulation is as follows,

\begin{equation}
\begin{aligned}
\label{eq:avg}
    g_i = D_{g_i}(x) \\
    g = \sum_{i=1}^{C} g_i \cdot w_i \\
    s.t. \sum_{i=1}^{C} w_i = 1
\end{aligned}
\end{equation}

where $g_i$ is the guidance vector generated from discriminator $D_{g_i}$ and $C$ represents the number of the discriminator, $w_i$ is computed via a neural network to learn the weights dynamically.

The formulation above requires the dimensions of the guidance vectors to be the same, which might be violated when there are heterogeneous architectures in different discriminators. It can be remedied by a module to map the guidances into the same embedding space. Another issue is the average guidance might be noisy because the directions of the guidance might diverge, which deteriorates the performance. The other approach is to use a memory-based neural network (i.e., bidirectional RNN \cite{pengMultiagentBidirectionallyCoordinatedNets2017a}) to address both issues. The hidden state in our recurrent network is a feature map instead of a vector, where noisy information exists. We adopt a bi-GRU architecture to utilize the gates to decrease unuseful information, which is formulated as follows:
\begin{equation}
\begin{aligned}
    g_i = NN_{GRU}(g_{i-1}; h_{i-1}) \\
    g = g_C
\end{aligned}
\end{equation}

where $g_i$ follows the definition in equation (\ref{eq:avg}) and $h_i$ is the hidden state of the bi-LSTM network.

After generating the guidance vector, it is piped into the generator to guide the image translation. We explore two types of merging methods. One is the classic merging, in which a neural module is employed to match the space of guidance and the generator's internal representation. Then another layer of a neural network merges the concatenation of two representations. In another setting, we explore using the self-attention mechanism, where a key-query-value triplet is constructed. The query is the guidance vector, the key is the generator latent representation, and the value is the merged representation.

\begin{equation}
\begin{aligned}
    z = G_{enc}(x) \\
    z_{new} = CONCAT(NN_{g}(g, z), z) \\
    z_{new} = CONCAT(Attention(z, g, g), z) \\
    Attention(K, Q, V) := Softmax(K^{T} Q) V \\
    \hat{x} = G_{dec}(z_{new})
\end{aligned}
\end{equation}

where $G_{enc}$ and $G_{dec}$ are the encoder and the decoder of the generator, the guidance is injected in the latent space, where $z$ is computed, $z_{new}$ is the latent representation after merging the guidance, $g$ is guidance, $K,Q,V$ represents key, query and value respectively.

\subsection{Regularization}
It is possible to add some regularization to the guidance vectors. Although there is no related study in the literature to the best our knowledge, the characteristics of the Image-to-Image translation makes it feasible to do so. In the translation process, we can add a new hypothesis that the guidance should reflect the saliency information about the image which can be modelled as the similarity constraints on the guidance. Specifically, we can add a simple $L_1$ norm on the guidance of true input image and the fake generated image.

\begin{equation}
\begin{aligned}
    L_{reg} =  \mathop{\mathbb{E}}_{x_A \sim p_A} || g_{x_A} - g_{x_{\hat{B}}} ||_1 \\
    + \mathop{\mathbb{E}}_{x_B \sim p_B} || g_{x_B} - g_{x_{\hat{A}}} ||_1
\end{aligned}
\end{equation}

\subsection{Full objective for generator training}
The full objective function of the generator training is as follows,
\begin{equation}
\begin{aligned}
L_{gen} = \lambda_{GAN} * L_{gan}(G) + \lambda_{cyc} * L_{cyc}  \\
    + \lambda_{reg} * L_{reg} \\
\end{aligned}
\end{equation}
where $L_{gan}$ is the GAN loss for generators which follows GAN loss with both direction $A \rightarrow B$ and $B \rightarrow A$, $L_{cyc}$ is the cycle consistency loss defined in the background description in section \ref{sec:methodology} and $L_{reg}$ is the regularization loss. We use hyperparameters $\lambda_{*}$ to control the impacts of different losses.

\subsection{Training for Discriminators}
In multi-agent RL, an extra communication module is added to compute the communication vectors. This module is trained jointly in each agent. Therefore, it is worth exploring whether the discriminator should be trained for guidance in the image translation task.

The goal is to send messages from discriminators to generators. One way is to use the discriminator parameters as data. Combined with the image information, the generator will be able to hack all the information contained in the environment. However, the computation complexity would be dependent with the size of discriminators. Furthermore, the discriminators do not provide additional information about the adversarial objective. We adopt the other way that use the discriminator to send processed continuous vectors to the generators. This communication design also enables discriminators to propagate useful information to generators given that the smaller size of message vectors form an information bottleneck.
Then there is gap that the training process of generators has a different objective with the discriminators'. From this point of view, when the discriminator is trained in the guidance, one model is used twice in the computation graph which will cause problems in training. Thus we introduce a two-step training algorithm for the discriminator, shown in the pseudo-code below.

\begin{algorithm}
\caption{Training for Discriminators}
\label{algo:training}
\begin{algorithmic}[1]
\STATE $\phi_D \gets $ initial parameters
\STATE $\phi_{\hat{D}} \gets \phi$

\REPEAT
\STATE $\phi_D \gets \phi_D - \alpha_{D} \nabla_{D}(- L_{gan})$
\STATE $\phi_{\hat{D}} \gets \phi_{\hat{D}} - \alpha_{\hat{D}} \nabla_{\hat{D}} L_{gen}$
\STATE $\phi_D \gets \lambda_D * \phi_D + (1 - \lambda_D) * \phi_{\hat{D}}$
\STATE $\phi_{\hat{D}} \gets \phi_D $
\UNTIL the $L_{cyc}$ converges \label{algo:training:until}

\end{algorithmic}
\end{algorithm}

In Algorithm \ref{algo:training}, $\phi_{\cdot}$ represents the trainable model parameters, $D$ is the discriminator without specification of domains, $\hat{D}$ is the corresponding copy of $D$. $\alpha$ is the learning rate for its relative model. We train the model iteratively with the GAN loss and the generator loss. Note in the GAN loss, the goal of the discriminator is to optimize the objective to its optimum. We use the batch gradient descent algorithm, so the objective turns into the negative version of itself. In the line \ref{algo:training:until}, the loop condition is set to the convergence of the cycle consistency, here we use the cycle-consistency as the index to measure the quality of the image generated. The GAN loss keeps flucatuation for the nature of the game. In other architectures, other criterion might be used to set the convergence condition.

\begin{figure*}
\centering

\includegraphics[scale = 0.22]{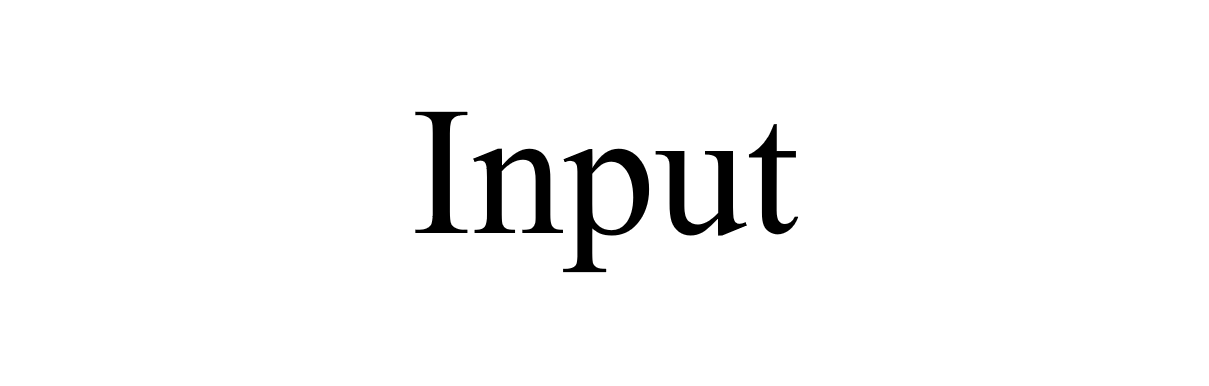}\enspace
\includegraphics[scale = 0.22]{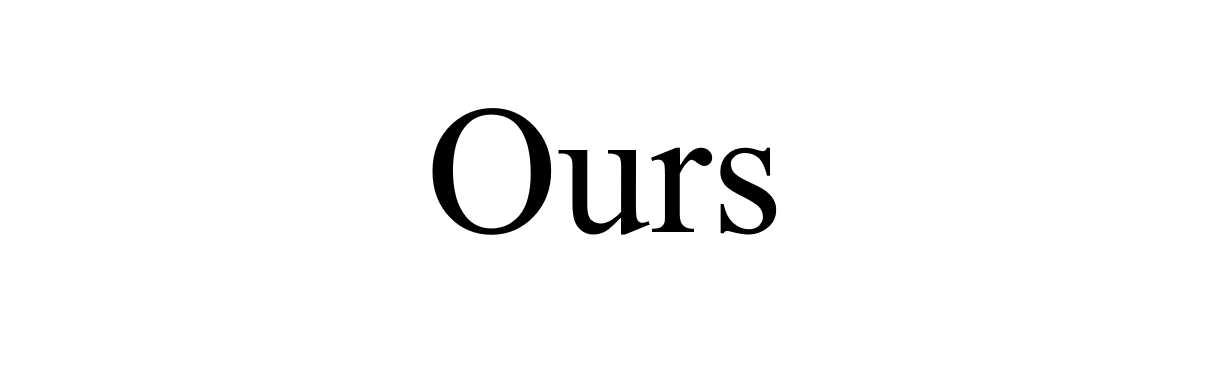}\enspace
\includegraphics[scale = 0.22]{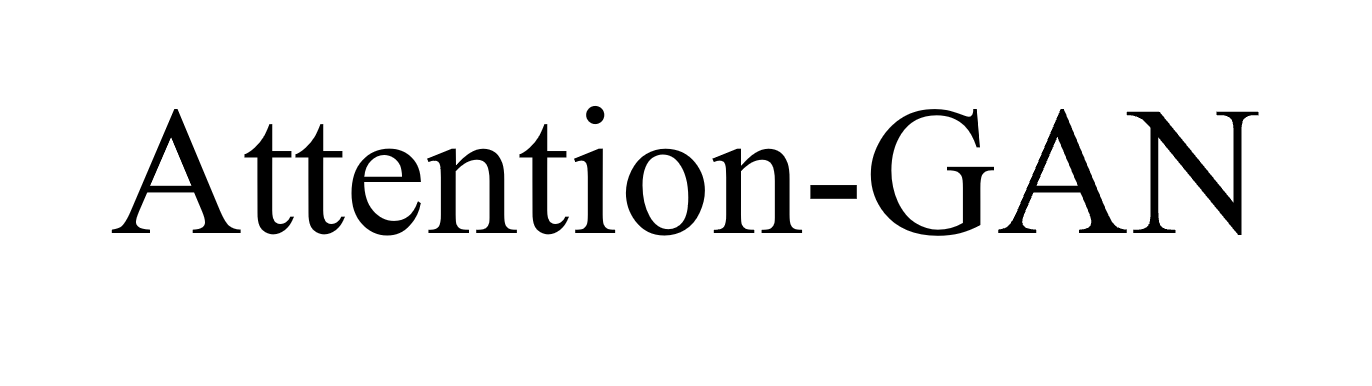}\enspace
\includegraphics[scale = 0.22]{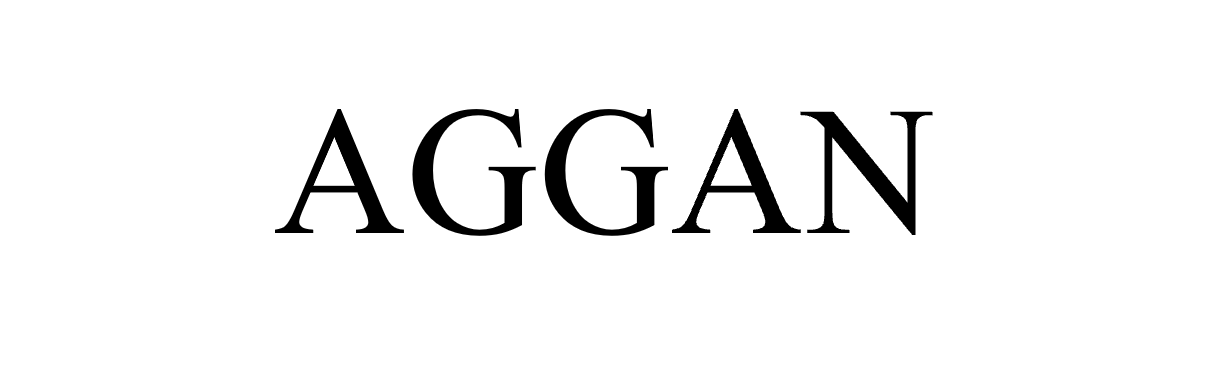}\enspace
\includegraphics[scale = 0.22]{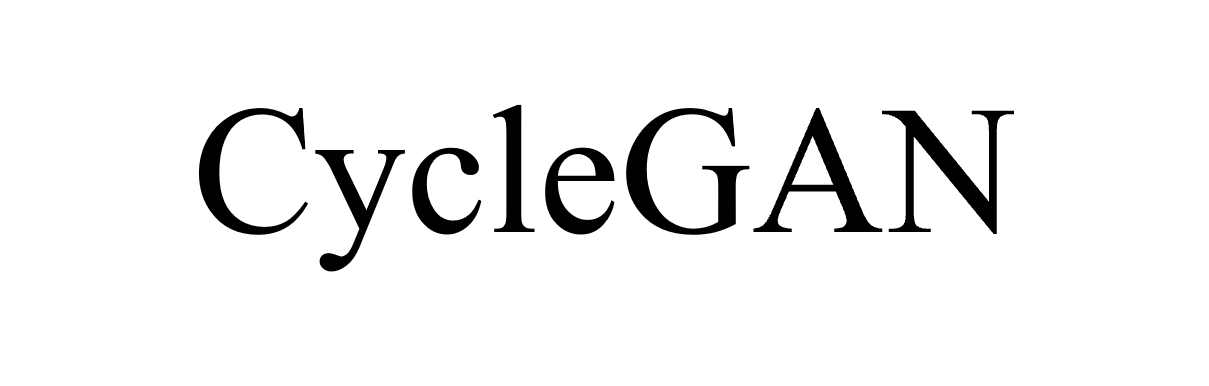}\enspace
\includegraphics[scale = 0.22]{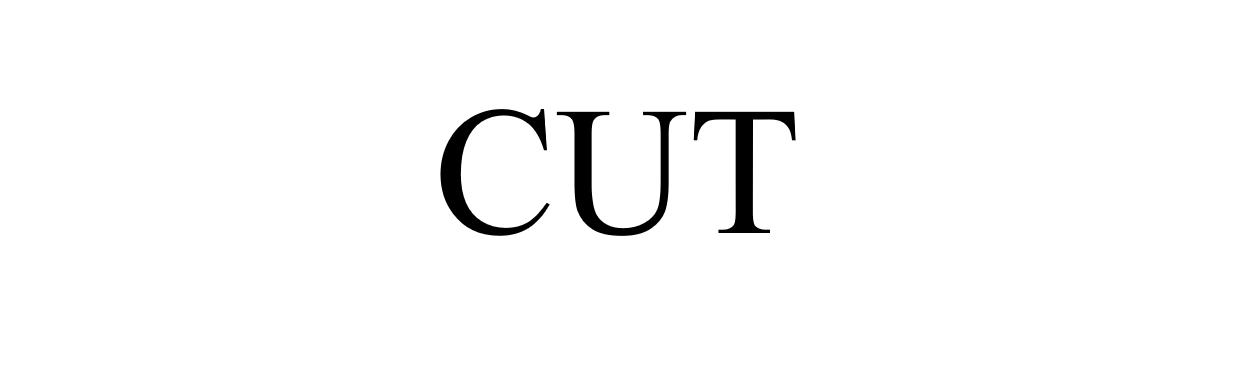}\enspace
\includegraphics[scale = 0.22]{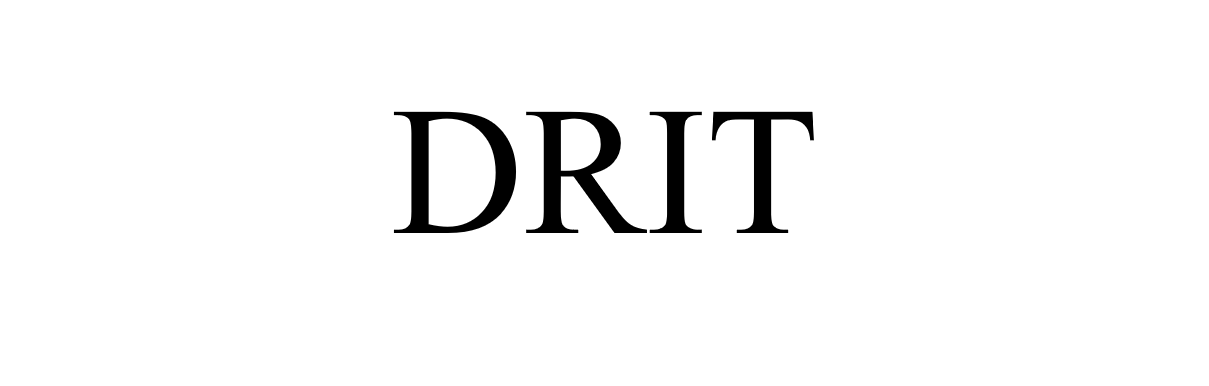}\enspace
\includegraphics[scale = 0.22]{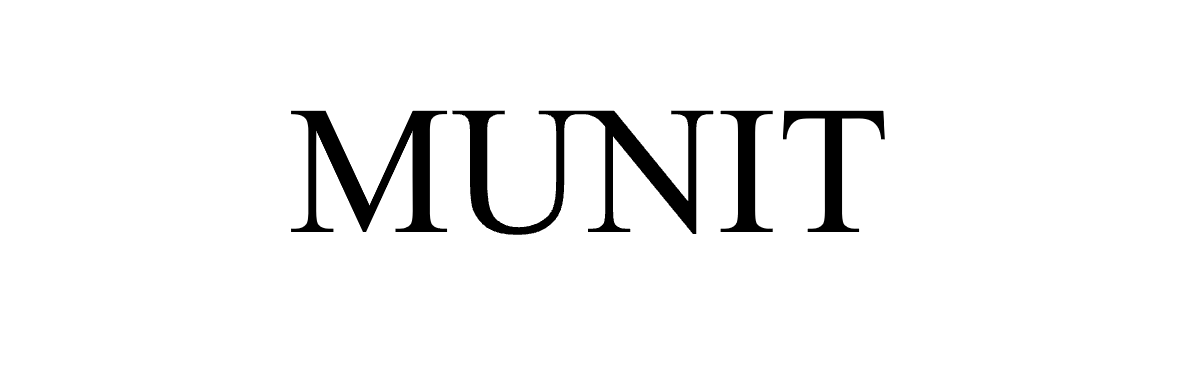}
\smallskip

\includegraphics[scale = 0.23]{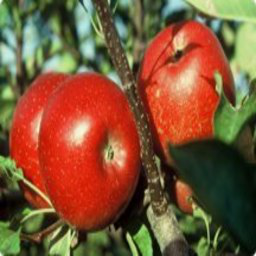}\enspace
\includegraphics[scale = 0.23]{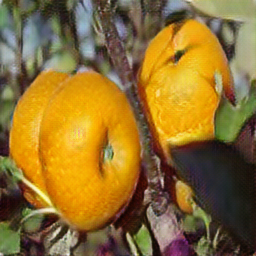}\enspace
\includegraphics[scale = 0.23]{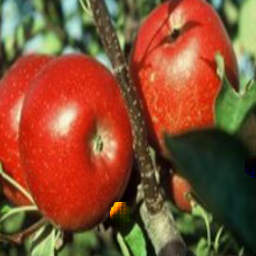}\enspace
\includegraphics[scale = 0.23]{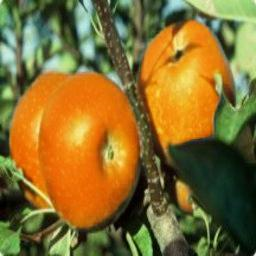}\enspace
\includegraphics[scale = 0.23]{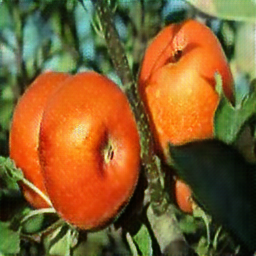}\enspace
\includegraphics[scale = 0.23]{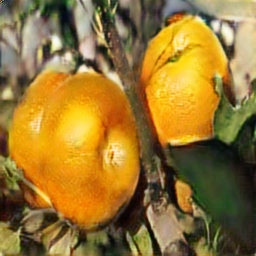}\enspace
\includegraphics[scale = 0.23]{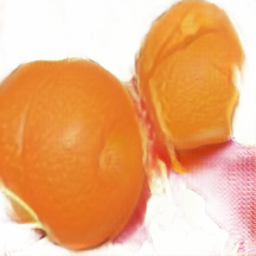}\enspace
\includegraphics[scale = 0.23]{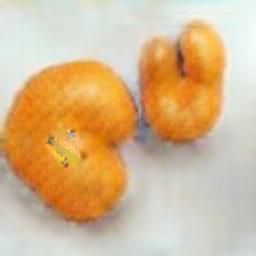}

\medskip

\includegraphics[scale = 0.23]{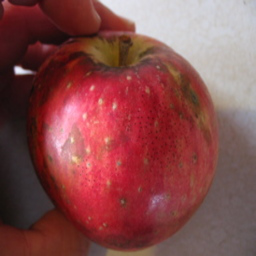}\enspace
\includegraphics[scale = 0.23]{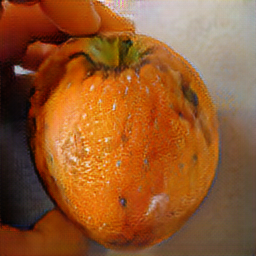}\enspace
\includegraphics[scale = 0.23]{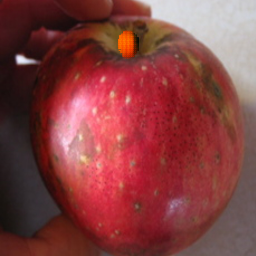}\enspace
\includegraphics[scale = 0.23]{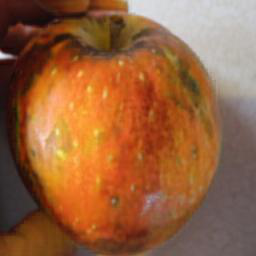}\enspace
\includegraphics[scale = 0.23]{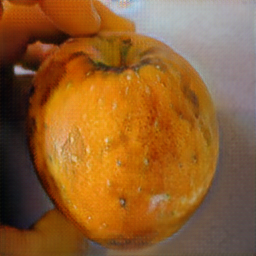}\enspace
\includegraphics[scale = 0.23]{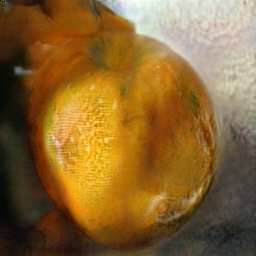}\enspace
\includegraphics[scale = 0.23]{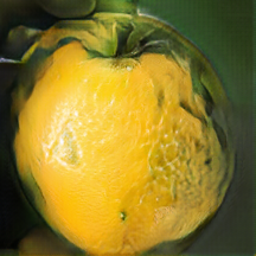}\enspace
\includegraphics[scale = 0.23]{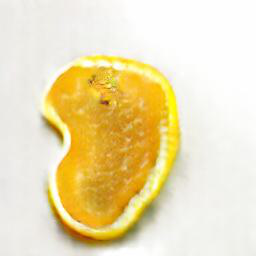}

\medskip

\includegraphics[scale = 0.23]{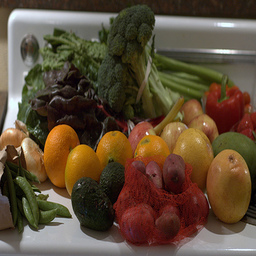}\enspace
\includegraphics[scale = 0.23]{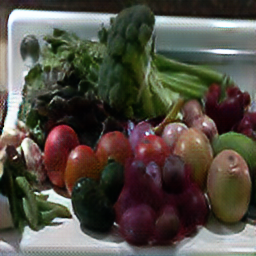}\enspace
\includegraphics[scale = 0.23]{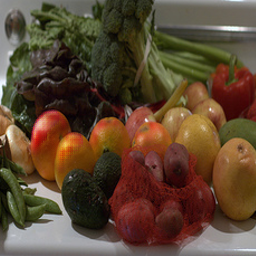}\enspace
\includegraphics[scale = 0.23]{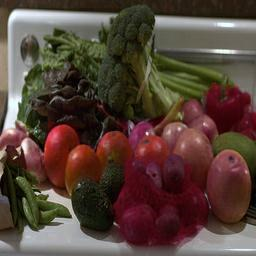}\enspace
\includegraphics[scale = 0.23]{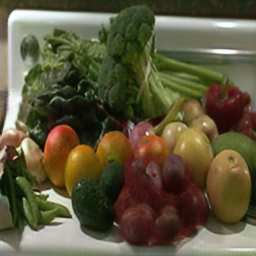}\enspace
\includegraphics[scale = 0.23]{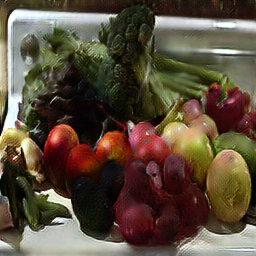}\enspace
\includegraphics[scale = 0.23]{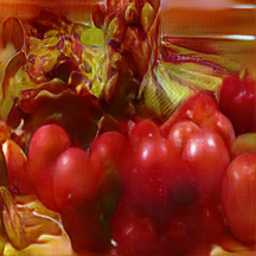}\enspace
\includegraphics[scale = 0.23]{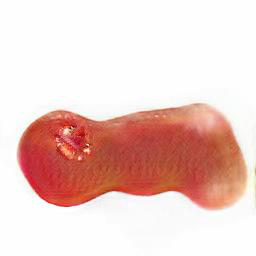}

\medskip

\includegraphics[scale = 0.23]{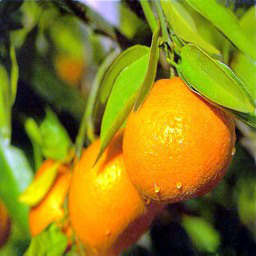}\enspace
\includegraphics[scale = 0.23]{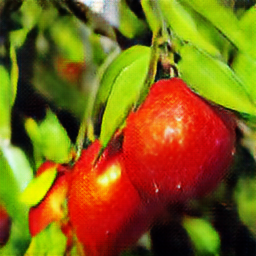}\enspace
\includegraphics[scale = 0.23]{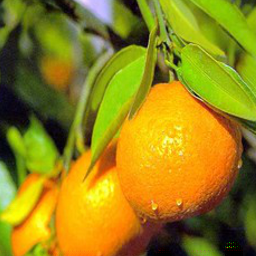}\enspace
\includegraphics[scale = 0.23]{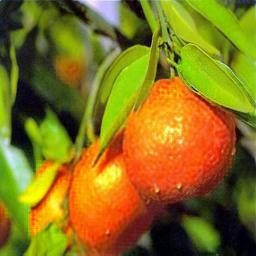}\enspace
\includegraphics[scale = 0.23]{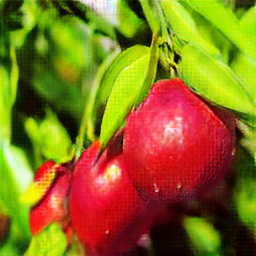}\enspace
\includegraphics[scale = 0.23]{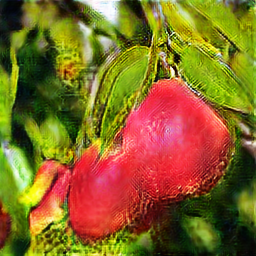}\enspace
\includegraphics[scale = 0.23]{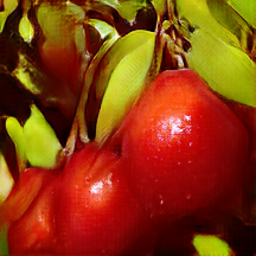}\enspace
\includegraphics[scale = 0.23]{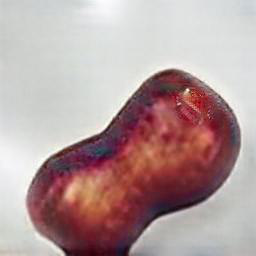}

\caption{Qualitative experiment results on apple2orange dataset, compared with multiple models}
\label{fig:apple}
\end{figure*}

\section{Experiments}
\label{sec:exp}

In this section, we will introduce our experiment setups, including dataset selection, comparison models and metrics. The results will be presented in a qualitative way with image examples as well as quantitative analysis by well adopted scores.

\subsection{Dataset and Environment}
In this section, we briefly introduce the datasets and the environment we used in the experiments:
\begin{itemize}
\item \textbf{Object transfiguration datasets} We evaluate our model on the widely used benchmark introduced by Zhu et al \cite{zhuUnpairedImagetoimageTranslation2017} where object transfiguration, season transfer and collection style transfer datasets are adopted. Our experiment aims at extracting objects from transfer learning. So we only adopt two object transfiguration datasets horse $\leftrightarrow$ zebra and apple $\leftrightarrow$ orange. Specifically, the apple $\leftrightarrow$ orange dataset contains 995/266 train/test apple images and 1019/248 train/test orange images; the horse $\leftrightarrow$ zebra dataset has 1067/120 train/test horse images and 1334/140 train/test zebra images.
Additionally, we also conduct experiments on the lion $\leftrightarrow$ tiger dataset used by \cite{emami2020spa}. The authors did not share their dataset, so we reproduce it by collecting images from ImageNet \cite{deng2009imagenet} in lion and tiger classes. Then we do the 4:1 train/test split on the datasets randomly. The lion $\leftrightarrow$ tiger dataset has 1683/403 train test tiger images and 1420/375 train/test lion images.

\item \textbf{Image preprocessing} To match the requirement of CycleGAN\cite{zhuUnpairedImagetoimageTranslation2017} backbone, we use the same preprocessing workflow for all images. The images are firstly scaled to $286 \times 286$, then cropped a $256 \times 256$ part as input. The input images are flipped with probability $0.5$.
\end{itemize}

\begin{figure*}
\centering

\includegraphics[scale = 0.22]{imgs/model-titles/input.png}\enspace
\includegraphics[scale = 0.22]{imgs/model-titles/ours.png}\enspace
\includegraphics[scale = 0.22]{imgs/model-titles/attngan.png}\enspace
\includegraphics[scale = 0.22]{imgs/model-titles/aggan.png}\enspace
\includegraphics[scale = 0.22]{imgs/model-titles/cycle.png}\enspace
\includegraphics[scale = 0.22]{imgs/model-titles/cut.png}\enspace
\includegraphics[scale = 0.22]{imgs/model-titles/drit.png}\enspace
\includegraphics[scale = 0.22]{imgs/model-titles/munit.png}
\smallskip

\includegraphics[scale = 0.23]{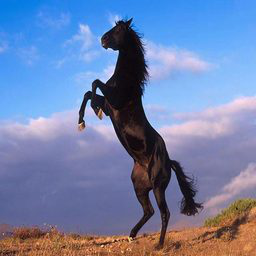}\enspace
\includegraphics[scale = 0.23]{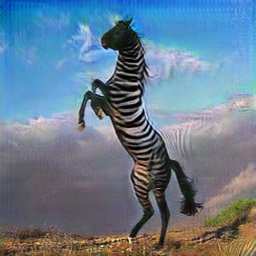}\enspace
\includegraphics[scale = 0.23]{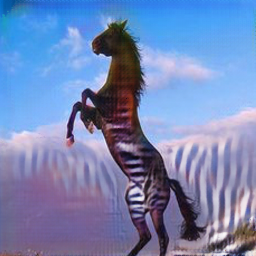}\enspace
\includegraphics[scale = 0.23]{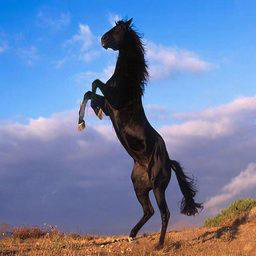}\enspace
\includegraphics[scale = 0.23]{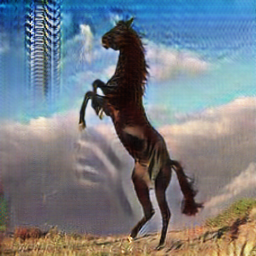}\enspace
\includegraphics[scale = 0.23]{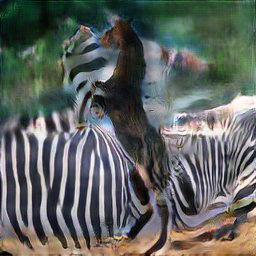}\enspace
\includegraphics[scale = 0.23]{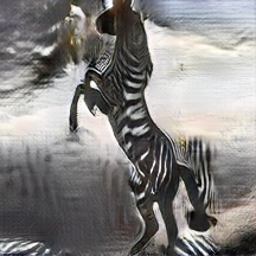}\enspace
\includegraphics[scale = 0.23]{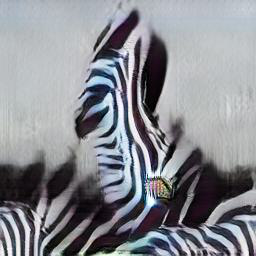}

\medskip

\includegraphics[scale = 0.23]{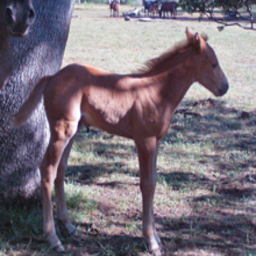}\enspace
\includegraphics[scale = 0.23]{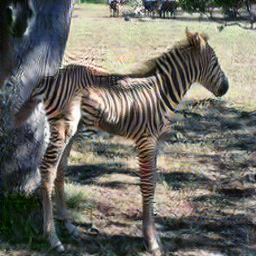}\enspace
\includegraphics[scale = 0.23]{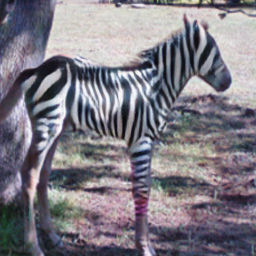}\enspace
\includegraphics[scale = 0.23]{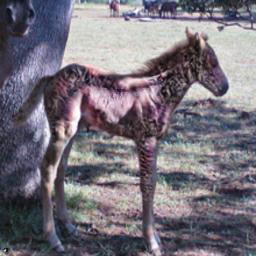}\enspace
\includegraphics[scale = 0.23]{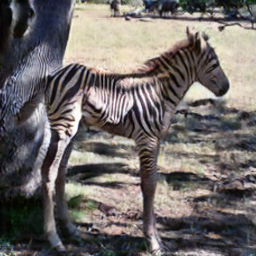}\enspace
\includegraphics[scale = 0.23]{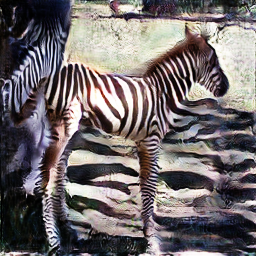}\enspace
\includegraphics[scale = 0.23]{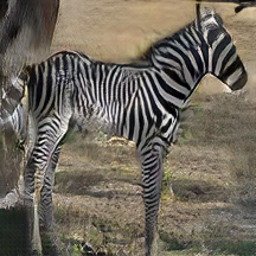}\enspace
\includegraphics[scale = 0.23]{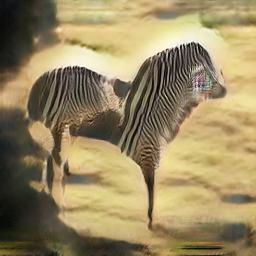}

\medskip

\includegraphics[scale = 0.23]{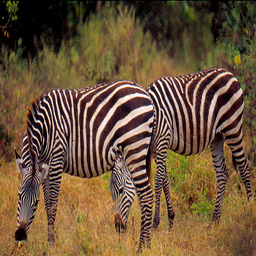}\enspace
\includegraphics[scale = 0.23]{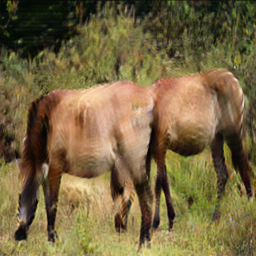}\enspace
\includegraphics[scale = 0.23]{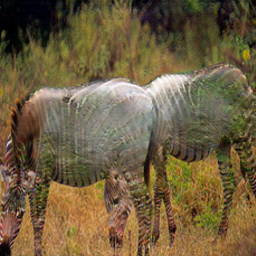}\enspace
\includegraphics[scale = 0.23]{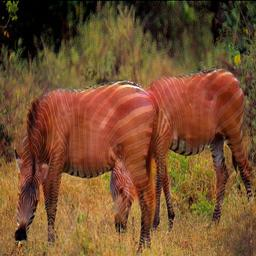}\enspace
\includegraphics[scale = 0.23]{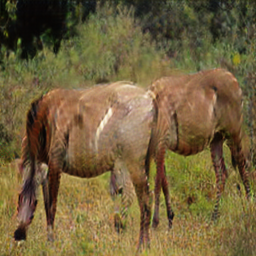}\enspace
\includegraphics[scale = 0.23]{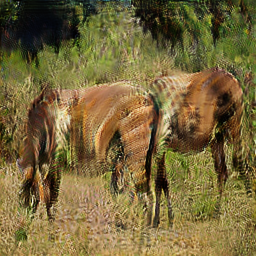}\enspace
\includegraphics[scale = 0.23]{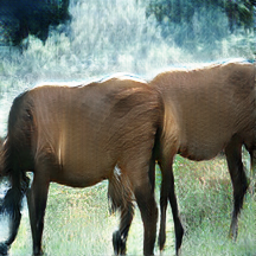}\enspace
\includegraphics[scale = 0.23]{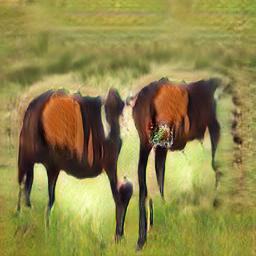}

\medskip

\includegraphics[scale = 0.23]{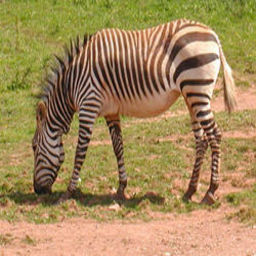}\enspace
\includegraphics[scale = 0.23]{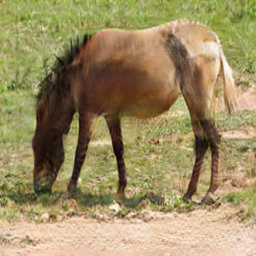}\enspace
\includegraphics[scale = 0.23]{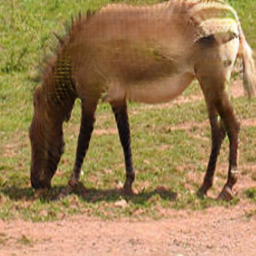}\enspace
\includegraphics[scale = 0.23]{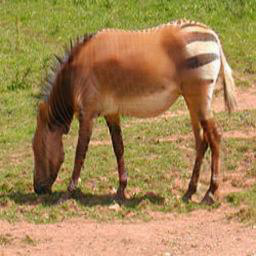}\enspace
\includegraphics[scale = 0.23]{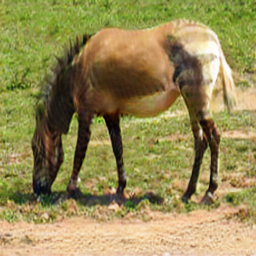}\enspace
\includegraphics[scale = 0.23]{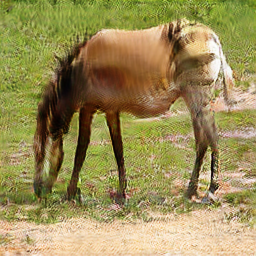}\enspace
\includegraphics[scale = 0.23]{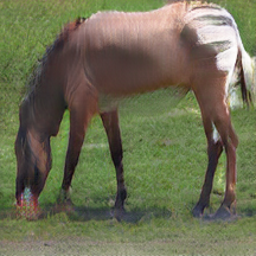}\enspace
\includegraphics[scale = 0.23]{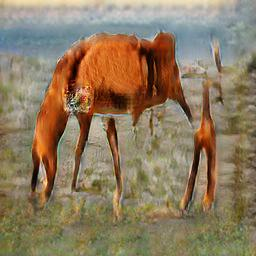}

\caption{Qualitative experiment results on horse2zebra dataset, compared with multiple models}
\label{fig:horse}
\end{figure*}

\begin{center} 
\begin{table*}
\caption{KID $\times 100 $ ± STD $\times 100$ results computed on only on target domain on three datasets. Smaller numbers are better.} 
\label{tab:model-comparison} 
\centering
\begin{tabular}{c|cccccc} 
\toprule
 Method & apple $\rightarrow$ orange & orange $\rightarrow$ apple & zebra $\rightarrow$ horse & horse $\rightarrow$ zebra & lion $\rightarrow$ tiger & tiger $\rightarrow$ lion \\
 \hline
 MUNIT & 13.45 ± 1.67 & 6.79 ± 0.78 & 6.32 ± 0.90 & 4.76 ± 0.63 & 2.67 ± 0.63 & 8.10 ± 0.87 \\
 DRIT & 9.65 ± 1.61 & 6.50 ± 1.16 & 5.67 ± 0.66 & 4.30 ± 0.57 & 2.39 ± 0.67 & 7.04 ± 0.73 \\
 CycleGAN & 11.02 ± 0.60 & 5.94 ± 0.65 & 4.87 ± 0.52 & 3.94 ± 0.41 & 2.56 ± 0.13 & 5.32 ± 0.47 \\
 AttentionGAN & 11.17 ± 0.92 & 5.41 ± 0.87 & 5.14 ± 0.68  & 4.67 ± 0.52 & 2.64 ± 0.24 & 6.28 ± 0.33 \\
 AGGAN &10.36 ± 0.86 & 4.54 ± 0.50 & 4.46 ± 0.40 & 4.12 ± 0.80 & 2.23 ± 0.21 & 5.83 ± 0.51 \\
 SPA-GAN &3.77 ± 0.32 & 2.38 ± 0.33 & 2.19 ± 0.12 & 2.01 ± 0.13 & 1.17 ± 0.19 & 3.09 ± 0.19 \\
 Our model & 6.21 ± 1.15 & 3.40 ± 0.49 & 2.57 ± 0.47 & 1.61 ± 0.52 & 1.51 ± 0.43 & 1.14 ± 0.55\\

\bottomrule
\end{tabular} 
\end{table*} 
\end{center}

\subsection{Compared methods}
We briefly introduce the comparison methods and metrics we used in this section.



\textbf{MUNIT}\cite{huangMultimodalUnsupervisedImagetoimage2018} decomposes the latent space into a content subspace and a style subspace. It translates an image by composing the content code with a style code from the other domain. In this way, they try to achieve multi-model translation.

\textbf{DRIT}\cite{leeDiverseImagetoimageTranslation2018} links disentangle representation into generation to combat mode collapse. The authors propose to use an adversarial loss to match the distribution in the content space.

\textbf{CUT}\cite{parkContrastiveLearningUnpaired2020} models the content by the contrastive InfoNCE loss. It treats corresponding patches across domains as positive samples and other parts as negative samples. The substitution of cycle-consistency enables the model to compute single direction translation only to boost training efficiency.

\textbf{CycleGAN}\cite{zhuUnpairedImagetoimageTranslation2017} introduces the widely used cycle-consistency loss that regularizes the fake images by the consistency loss between double generated images and original images. This strong regularization dramatically improves the similarity between the generated images and the input images.

\textbf{AttentionGAN}\cite{chenAttentionganObjectTransfiguration2018} generates fake images and attention maps separately in two neural networks. The attention maps that capture the region of interest in each domain are trained by multiplying them by original images and generated images. It enhances the performance by a consistency loss for attention maps.

\textbf{AGGAN}\cite{mejjati2018unsupervised} generates the attention maps by generator and combines them by product operation like AttentionGAN. But it considers the relationship between the attention maps and the discriminators, and it multiplies the attention maps with the original images into discriminators. The attention maps are discretized by a threshold to filter out noisy pixels with small probability values. The discriminator regularizes the maps during the training.

\textbf{SpaGAN}\cite{emami2020spa} generates attention maps from discriminator feature maps that improve the computation efficiency(AttentionGAN and AGGAN all generate attention maps in generators). It uses attention maps to guide the generator for better performance.

We select the Kernel Inception Distance(KID) \cite{binkowski2018demystifying} as the comparison metric. It is defined as the squared MMD distance between Inception representation vectors for images, which can be viewed as a kernel representation distance on images. Other metrics like Inception Score and FID got their disadvantages. The Inception Score relies on the KL divergence between the prior and posterior distribution of classes. When the images are sampled from classes out of Imagenet 1000 classes, it cannot capture the distribution of the images. For FID, \cite{binkowski2018demystifying}'s experiments show that the variance of estimates is still unignorable even in the presence of a large number of data samples. Even for the same amount of dataset size, examples are discovered that FID might be various. Also, FID hypothesizes that the distribution of the activations is Gaussian, which poses extra limitations compared to KID. Prior works \cite{emami2020spa, cazenavette2021mixergan} adopt KID as a comparison metric. We evaluate our model with both two KID computation methods. One is mentioned in \cite{mejjati2018unsupervised}, and the other is in \cite{emami2020spa}. The first one considers the background similarity by computing KID with both the source and target domain. The other removes the input domain because some of the backgrounds in the dataset are meaningless, such as blank backgrounds in the apple $\leftrightarrow$ orange dataset.

\subsection{Implementation details}
We intialize the learning rate at $2 \times 10^{-4}$ and configure the learning rate to decay linearly (timed by $0.1$) every 100 epochs. The whole process is optimized by Adam optimizer~\cite{kingma2014adam}, whose hyper-parameters $\beta_1=0.5, \beta_2=0.999$. The model is trained in the batch size of 1. Following~\cite{zhuUnpairedImagetoimageTranslation2017}, we use an image pool for discriminator training to randomly sample data for discriminators, roughly describing the distribution of the shifting distribution of real and generated samples. As for model implementation, we follow~\cite{zhuUnpairedImagetoimageTranslation2017} and employ the LSGAN architecture, where generators are based on ResNet~\cite{he2016DeepResidual}. We modify the output layer of the model to compute the loss we listed in Section \ref{sec:methodology}.
We adopt a two-layer discriminator implementation on some datasets.

\begin{center} 
\begin{table*}
\caption{KID $\times 100 $ ± STD $\times 100$ results computed on both source and target domains on three datasets. Smaller numbers are better.} 
\label{tab:model-comparison-two-domain} 
\centering
\begin{tabular}{c|cccccc} 
\toprule
 Method & apple $\rightarrow$ orange & orange $\rightarrow$ apple & zebra $\rightarrow$ horse & horse $\rightarrow$ zebra & lion $\rightarrow$ tiger & tiger $\rightarrow$ lion \\
 \hline
 MUNIT & 9.70 ± 1.22 & 10.61 ± 1.16 & 11.51 ± 1.27 & 8.31 ± 0.46 & 10.87 ± 0.91 & 10.61 ± 0.47 \\
 DRIT & 6.37 ± 0.75 & 8.34 ± 1.22 & 9.65 ± 0.91 & 8.23 ± 0.08 & 9.56 ± 0.18 & 10.11 ± 0.59 \\
 CycleGAN & 8.48 ± 0.53 & 9.82 ± 0.51 & 11.44 ± 0.38 & 10.25 ± 0.25 & 10.15 ± 0.08 & 10.97 ± 0.04 \\
 AttentionGAN & 7.90 ± 0.25 & 8.05 ± 0.49 & 9.86 ± 0.32 & 8.28 ± 0.34 & 10.35 ± 0.58 & 10.56 ± 0.65 \\
 AGGAN & 6.44 ± 0.69 & 5.32 ± 0.48 & 8.87 ± 0.26 & 6.93 ± 0.27 & 8.56 ± 0.16 & 9.17 ± 0.07 \\
 SPA-GAN & 5.81 ± 0.51 & 7.95 ± 0.42 & 8.72 ± 0.24 & 7.89 ± 0.29 & 8.47 ± 0.07 & 8.63 ± 0.05\\
 Our model & 5.21 ± 0.72 & 5.29 ± 1.18 & 9.55 ± 1.03 & 7.81 ± 0.66 & 8.18 ± 1.04 & 8.24 ± 0.96  \\

\bottomrule
\end{tabular} 
\end{table*} 
\end{center}

\subsection{Results}
In this section, we list the results of our experiments, which includes quantitative analysis and qualitative analysis.

\subsubsection{Qualitative analysis for Image Translation}
\begin{figure*}
\centering

\includegraphics[scale = 0.22]{imgs/model-titles/input.png}\enspace
\includegraphics[scale = 0.22]{imgs/model-titles/ours.png}\enspace
\includegraphics[scale = 0.22]{imgs/model-titles/attngan.png}\enspace
\includegraphics[scale = 0.22]{imgs/model-titles/aggan.png}\enspace
\includegraphics[scale = 0.22]{imgs/model-titles/cycle.png}\enspace
\includegraphics[scale = 0.22]{imgs/model-titles/cut.png}\enspace
\includegraphics[scale = 0.22]{imgs/model-titles/drit.png}\enspace
\includegraphics[scale = 0.22]{imgs/model-titles/munit.png}
\smallskip

\includegraphics[scale = 0.23]{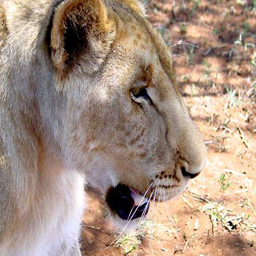}\enspace
\includegraphics[scale = 0.23]{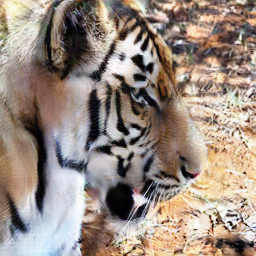}\enspace
\includegraphics[scale = 0.23]{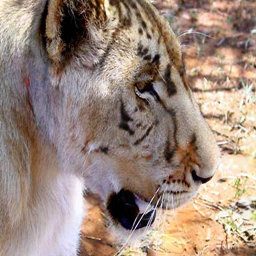}\enspace
\includegraphics[scale = 0.23]{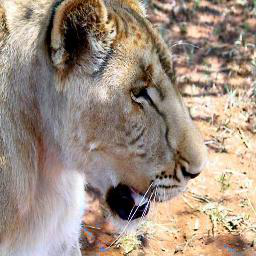}\enspace
\includegraphics[scale = 0.23]{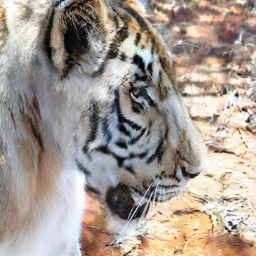}\enspace
\includegraphics[scale = 0.23]{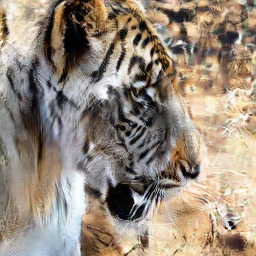}\enspace
\includegraphics[scale = 0.23]{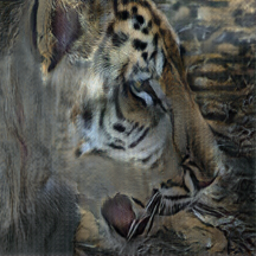}\enspace
\includegraphics[scale = 0.23]{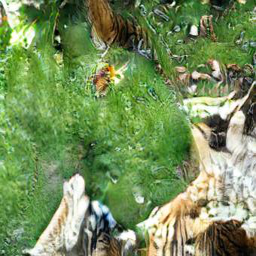}

\medskip

\includegraphics[scale = 0.23]{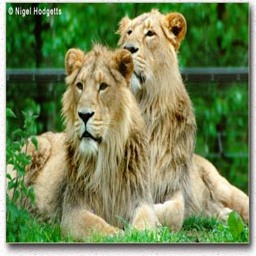}\enspace
\includegraphics[scale = 0.23]{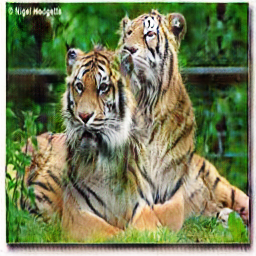}\enspace
\includegraphics[scale = 0.23]{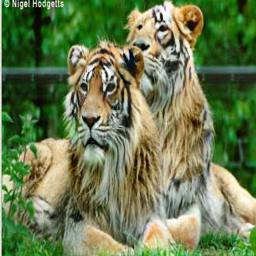}\enspace
\includegraphics[scale = 0.23]{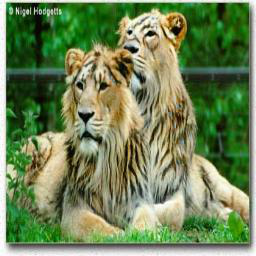}\enspace
\includegraphics[scale = 0.23]{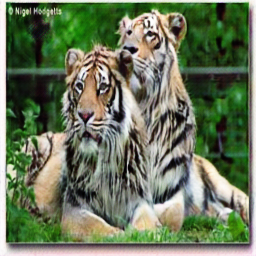}\enspace
\includegraphics[scale = 0.23]{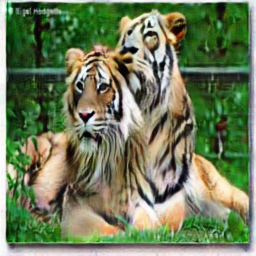}\enspace
\includegraphics[scale = 0.23]{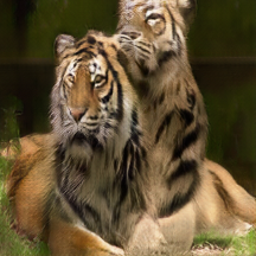}\enspace
\includegraphics[scale = 0.23]{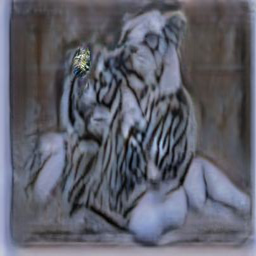}

\medskip

\includegraphics[scale = 0.23]{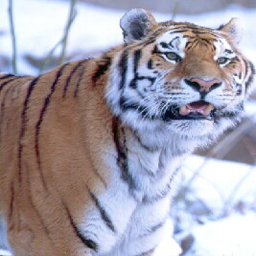}\enspace
\includegraphics[scale = 0.23]{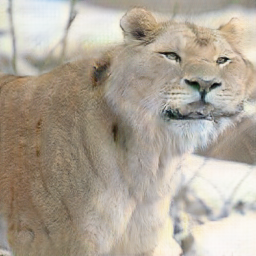}\enspace
\includegraphics[scale = 0.23]{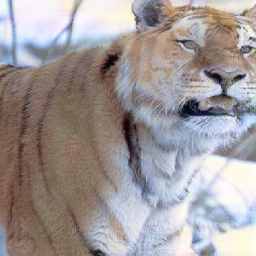}\enspace
\includegraphics[scale = 0.23]{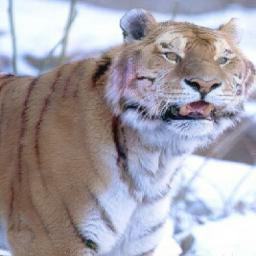}\enspace
\includegraphics[scale = 0.23]{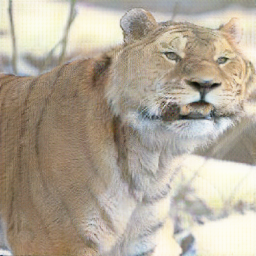}\enspace
\includegraphics[scale = 0.23]{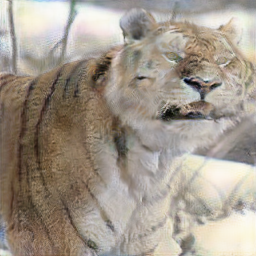}\enspace
\includegraphics[scale = 0.23]{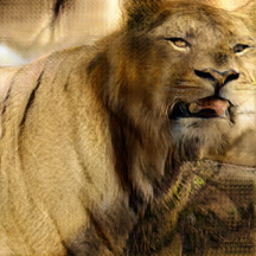}\enspace
\includegraphics[scale = 0.23]{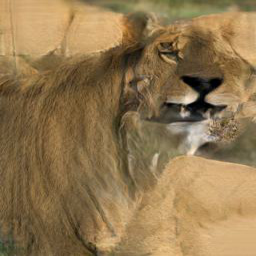}

\medskip

\includegraphics[scale = 0.23]{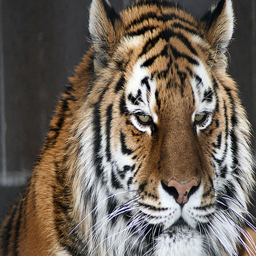}\enspace
\includegraphics[scale = 0.23]{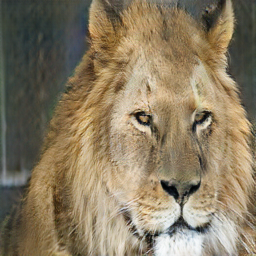}\enspace
\includegraphics[scale = 0.23]{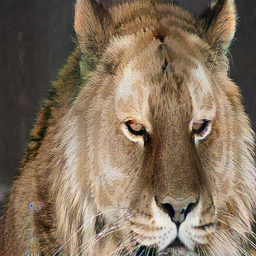}\enspace
\includegraphics[scale = 0.23]{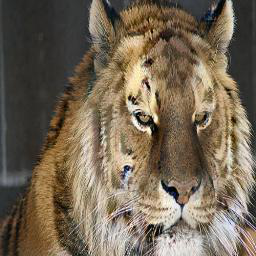}\enspace
\includegraphics[scale = 0.23]{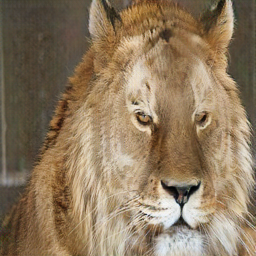}\enspace
\includegraphics[scale = 0.23]{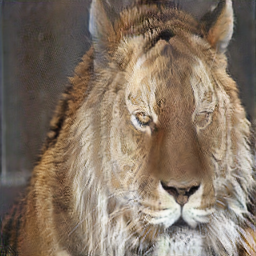}\enspace
\includegraphics[scale = 0.23]{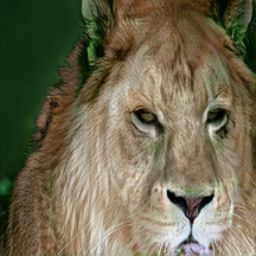}\enspace
\includegraphics[scale = 0.23]{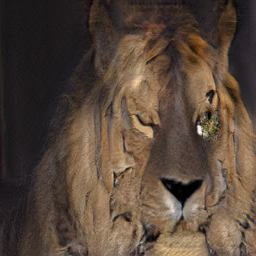}

\caption{Qualitative experiment results on lion2tiger dataset, compared with multiple models}
\label{fig:lion}
\end{figure*}

We demonstrate the performance of our model by qualitative analysis in which typical images are sampled for the comparison of our model and others. The results are listed in the Figure \ref{fig:apple}, Figure \ref{fig:horse} and Figure \ref{fig:lion}. 

In Figure \ref{fig:apple}, the first two rows of images show the translation from the apple domain to the orange domain and the last two rows demonstrate the reverse translation. Our model is better in the sense that it can capture all the textures to be transformed, and it can generate better textures. For example, in row two, AttentionGAN fails to capture the apple texture with green dots. CUT, DRIT and MUNIT fail to keep the background; either the fingers are blurred or just disappear. CycleGAN and AGGAN have good results, but the texture quality is still worse than ours. In the orange translation, our model can cover all the transformation parts in the images. In the third row, all oranges are transformed. In row 4, oranges in the background are captured and translated with the similar red color. In Figure \ref{fig:horse}, we perform a good transformation with little background twisted in the first two rows. Note that in row 2, the eyes of horses and shadows well-kept without explicit masks in our model's result. The last row shows that we can cover the difficult large stripes on zebras. Figure \ref{fig:lion} shows that our model can produce high-quality texture from lion to tiger shown in the first row, including transforming lions' mane into well-shaped stripes. The reverse transformation shows that our model can produce higher quality lion images by removing or fading the stripes, especially the stripes on heads.
We don't do the qualitative analysis on the SPA-GAN model because the source code of the model is not released. 

\subsubsection{Quantitative analysis by Kernel Inception Distance}
We show our model's performance in the form of distribution discrepancy in Table \ref{tab:model-comparison}. Our model can achieve the best performance in some domains and the second-best in all other domains. Table \ref{tab:model-comparison} shows the result of single domain translation, where the KID is computed between generated fake images and real images on target domain. Table \ref{tab:model-comparison-two-domain} shows the KID computed between generated fake images and both two domains. It can be seen that our model achieves the best performance on two datasets.

\subsection{Analysis expriment details}

The training of GAN is time-consuming, with limited computation resources, we choose to reduce the dataset during the training process for the analysis part. We adopt the apple $\leftrightarrow$ orange dataset for analysis because its size is small and the large variance of performance between can better show the analysis results, according to the KID in table \ref{tab:model-comparison}

We cut the size to 1/5 of its original size. To prevent a large performance drop from the real dataset to the reduced compact dataset, we use a simple subsample method. Firstly, we use a ResNet model trained on ImageNet \cite{deng2009imagenet} to map each image to an embedding vector. Then we compute the mean of the embedding dataset, and sort it by the distance between each embedding vector to the mean. Set the size of the dataset is $N$. Then we separate the sorted embedding vector into $\lfloor N/5 \rfloor$ number of blocks, we sample one image in each block and form a new dataset. Experiments show that the model trained on the subsampled dataset has a tolerable performance gap to a model trained on the real dataset.

\subsection{Ablation Study}

\begin{center} 
\begin{table}
\caption{KID $\times 100 $ ± STD $\times 100$ results computed for Ablation study. Smaller numbers are better.}
\label{tab:ablation} 
\centering
\begin{tabular}{c|cc} 
\toprule
 Method & apple $\rightarrow$ orange & orange $\rightarrow$ apple \\
 \hline
CycleGAN & 13.86 ± 1.02 & 5.95 ± 0.66 \\
+ guidance & 13.23 ± 1.79 & 5.20 ± 0.57\\
+ attention based guidance & 11.67 ± 1.93 & 4.47 ± 0.75 \\
+ attention guidance + regularization & 10.82 ± 1.09 & 5.96 ± 0.67 \\
+ multi-attention-guidance & 10.52 ± 1.37 & 5.59 ± 0.67 \\
\bottomrule
\end{tabular} 
\end{table}
\end{center}

To validate the effectiveness of our model, we conduct ablation experiments on the apple dataset. The KID of the CycleGAN model is computed on a locally trained reproduction model. As shown in Table \ref{tab:ablation}, we compared models mentioned in Section \ref{sec:methodology}. The table illustrates that the guidance module is effective. When the guidance is simply a convolution fusion layer, the result will be imbalanced. When attention-based guidance is employed, the result becomes better. We also compare the performance of regularization and multi-guidance schemes. When regularization is applied, the performance moves slightly towards the balanced result. Compared to single guidance, the multi-guidance model has worse performance. We suspect that this is due to the multi-guidance cannot provide more information than a single model but it indeed increases computation complexity where the gradient is hard to propagate through. We show the empirical results support this hypothesis in Table \ref{tab:multi-guidance}.

\subsection{Single discriminator exploration}

\begin{center} 
\begin{table}
\caption{KID $\times 100 $ ± STD $\times 100$ results computed for domain guidance analysis.}
\label{tab:single-guidance} 
\centering
\begin{tabular}{c|cc} 
\toprule
 Method & apple $\rightarrow$ orange & orange $\rightarrow$ apple \\
 \hline
CycleGAN & 13.86 ± 1.02 & 5.95 ± 0.66 \\
input domain guidance & 10.17 ± 1.69 & 4.95 ± 0.50  \\
output domain guidance & 11.67 ± 1.93 & 4.47 ± 0.75 \\
\bottomrule
\end{tabular} 
\end{table}
\end{center}

In this section, we conduct experiments on single discriminator guidance. As mentioned in section \ref{sec:methodology}, we classify guidance into two classes, input domain guidance and output domain guidance. The input and output are defined according to the generator. Each domain provides useful information for the generator. And we discuss the result of the experiments and some interesting insights.


Our experiments show that both the guidance improves the generator performance. One interesting observation is that the output domain guidance increases the model performance without changing the imbalance between domains, but the input domain guidance does. Furthermore, when we impose the guidance constraint, the imbalance reappears in the guidance again.

\subsection{Merging strategy exploration}

\begin{center} 
\begin{table}
\caption{KID $\times 100 $ ± STD $\times 100$ results computed for multi-communication analysis.} 
\label{tab:multi-guidance} 
\centering
\begin{tabular}{c|cc} 
\toprule
 Method & apple $\rightarrow$ orange & orange $\rightarrow$ apple \\
 \hline
CycleGAN & 13.86 ± 1.02 & 5.95 ± 0.66 \\
average & 10.52 ± 1.37 & 5.59 ± 0.67 \\
weighted average & 14.91 ± 2.03 & 7.77 ± 1.00 \\
bi-GRU & 13.31 ± 1.51 & 5.39 ± 0.72 \\
\bottomrule
\end{tabular} 
\end{table} 
\end{center}

We analyze the reason behind the performance drop by comparing different performances of the multi-guidance model. As described in section \ref{sec:methodology}, we explore several fusion strategies. The results are placed in Table \ref{tab:multi-guidance}. The performance of CycleGAN is used as the baseline. It can be seen from the table that with the new information provided by the guidance, the numbers of the multi-guidance model are higher than that of CycleGAN. While data fusion by simple average has almost achieved better performance. The hierarchical attention-based weighted average and bi-GRU have not made much increase. The performance bottleneck can be explained by the fact that if we simply add the weighted average without a shortcut, the performance might be worse.

\section{Conclusion}
In this work, we explore the communication mechanism between discriminators and generators in the GAN architecture for unpaired I2I translation. We formulate GAN as a Partially-observed Markov Decision Process (POMDP) for generators and propose adding a communication channel between discriminators and generators, which is inspired by multi-agent Reinforcement Learning. We conduct quantitative and qualitative analyses to evaluate our model, and  the results empirically show the effectiveness of our proposed approach.




 
\bibliographystyle{IEEEtran}
\bibliography{mybib}

\begin{thebibliography}{10}
\providecommand{\url}[1]{#1}
\csname url@samestyle\endcsname
\providecommand{\newblock}{\relax}
\providecommand{\bibinfo}[2]{#2}
\providecommand{\BIBentrySTDinterwordspacing}{\spaceskip=0pt\relax}
\providecommand{\BIBentryALTinterwordstretchfactor}{4}
\providecommand{\BIBentryALTinterwordspacing}{\spaceskip=\fontdimen2\font plus
\BIBentryALTinterwordstretchfactor\fontdimen3\font minus
  \fontdimen4\font\relax}
\providecommand{\BIBforeignlanguage}[2]{{%
\expandafter\ifx\csname l@#1\endcsname\relax
\typeout{** WARNING: IEEEtran.bst: No hyphenation pattern has been}%
\typeout{** loaded for the language `#1'. Using the pattern for}%
\typeout{** the default language instead.}%
\else
\language=\csname l@#1\endcsname
\fi
#2}}
\providecommand{\BIBdecl}{\relax}
\BIBdecl

\bibitem{zheng2021one}
Z.~Zheng, Z.~Yu, H.~Zheng, Y.~Yang, and H.~T. Shen, ``One-shot image-to-image
  translation via part-global learning with a multi-adversarial framework,''
  \emph{IEEE Transactions on Multimedia}, vol.~24, pp. 480--491, 2021.

\bibitem{zheng2022asynchronous}
Z.~Zheng, Y.~Bin, X.~Lu, Y.~Wu, Y.~Yang, and H.~T. Shen, ``Asynchronous
  generative adversarial network for asymmetric unpaired image-to-image
  translation,'' \emph{IEEE Transactions on Multimedia}, 2022.

\bibitem{chen2019quality}
L.~Chen, L.~Wu, Z.~Hu, and M.~Wang, ``Quality-aware unpaired image-to-image
  translation,'' \emph{IEEE Transactions on Multimedia}, vol.~21, no.~10, pp.
  2664--2674, 2019.

\bibitem{pang2021image}
Y.~Pang, J.~Lin, T.~Qin, and Z.~Chen, ``Image-to-image translation: Methods and
  applications,'' \emph{IEEE Transactions on Multimedia}, vol.~24, pp.
  3859--3881, 2021.

\bibitem{huang2021multi}
J.~Huang, L.~Jing, Z.~Tan, and S.~Kwong, ``Multi-density sketch-to-image
  translation network,'' \emph{IEEE Transactions on Multimedia}, vol.~24, pp.
  4002--4015, 2021.

\bibitem{kimUgatitUnsupervisedGenerative2019}
J.~Kim, M.~Kim, H.~Kang, and K.~Lee, ``U-gat-it: {{Unsupervised}} generative
  attentional networks with adaptive layer-instance normalization for
  image-to-image translation,'' \emph{arXiv preprint arXiv:1907.10830}, 2019.

\bibitem{dong2015image}
C.~Dong, C.~C. Loy, K.~He, and X.~Tang, ``Image super-resolution using deep
  convolutional networks,'' \emph{IEEE transactions on pattern analysis and
  machine intelligence}, vol.~38, no.~2, pp. 295--307, 2015.

\bibitem{gatysImageStyleTransfer2016a}
L.~A. Gatys, A.~S. Ecker, and M.~Bethge, ``Image style transfer using
  convolutional neural networks,'' in \emph{Proceedings of the {{IEEE}}
  Conference on Computer Vision and Pattern Recognition}, 2016, pp. 2414--2423.

\bibitem{guo2020gan}
X.~Guo, Z.~Wang, Q.~Yang, W.~Lv, X.~Liu, Q.~Wu, and J.~Huang, ``Gan-based
  virtual-to-real image translation for urban scene semantic segmentation,''
  \emph{Neurocomputing}, vol. 394, pp. 127--135, 2020.

\bibitem{rao2020rl}
K.~Rao, C.~Harris, A.~Irpan, S.~Levine, J.~Ibarz, and M.~Khansari,
  ``Rl-cyclegan: Reinforcement learning aware simulation-to-real,'' in
  \emph{Proceedings of the IEEE/CVF Conference on Computer Vision and Pattern
  Recognition}, 2020, pp. 11\,157--11\,166.

\bibitem{huangMultimodalUnsupervisedImagetoimage2018}
X.~Huang, M.-Y. Liu, S.~Belongie, and J.~Kautz, ``Multimodal unsupervised
  image-to-image translation,'' in \emph{Proceedings of the {{European}}
  Conference on Computer Vision ({{ECCV}})}, 2018, pp. 172--189.

\bibitem{zhuUnpairedImagetoimageTranslation2017}
J.-Y. Zhu, T.~Park, P.~Isola, and A.~A. Efros, ``Unpaired image-to-image
  translation using cycle-consistent adversarial networks,'' in
  \emph{Proceedings of the {{IEEE}} International Conference on Computer
  Vision}, 2017, pp. 2223--2232.

\bibitem{goodfellow2020generative}
I.~Goodfellow, J.~Pouget-Abadie, M.~Mirza, B.~Xu, D.~Warde-Farley, S.~Ozair,
  A.~Courville, and Y.~Bengio, ``Generative adversarial networks,''
  \emph{Communications of the ACM}, vol.~63, no.~11, pp. 139--144, 2020.

\bibitem{gronauerMultiagentDeepReinforcement2022a}
S.~Gronauer and K.~Diepold, ``Multi-agent deep reinforcement learning: A
  survey,'' \emph{Artificial Intelligence Review}, vol.~55, no.~2, pp.
  895--943, Feb. 2022.

\bibitem{mejjati2018unsupervised}
\BIBentryALTinterwordspacing
Y.~Alami~Mejjati, C.~Richardt, J.~Tompkin, D.~Cosker, and K.~I. Kim,
  ``Unsupervised attention-guided image-to-image translation,'' in
  \emph{Advances in Neural Information Processing Systems}, S.~Bengio,
  H.~Wallach, H.~Larochelle, K.~Grauman, N.~Cesa-Bianchi, and R.~Garnett, Eds.,
  vol.~31.\hskip 1em plus 0.5em minus 0.4em\relax Curran Associates, Inc.,
  2018. [Online]. Available:
  \url{https://proceedings.neurips.cc/paper/2018/file/4e87337f366f72daa424dae11df0538c-Paper.pdf}
\BIBentrySTDinterwordspacing

\bibitem{emami2020spa}
H.~Emami, M.~M. Aliabadi, M.~Dong, and R.~B. Chinnam, ``Spa-gan: Spatial
  attention gan for image-to-image translation,'' \emph{IEEE Transactions on
  Multimedia}, vol.~23, pp. 391--401, 2020.

\bibitem{isolaImagetoimageTranslationConditional2017}
P.~Isola, J.-Y. Zhu, T.~Zhou, and A.~A. Efros, ``Image-to-image translation
  with conditional adversarial networks,'' in \emph{Proceedings of the {{IEEE}}
  Conference on Computer Vision and Pattern Recognition}, 2017, pp. 1125--1134.

\bibitem{chenAttentionganObjectTransfiguration2018}
X.~Chen, C.~Xu, X.~Yang, and D.~Tao, ``Attention-gan for object transfiguration
  in wild images,'' in \emph{Proceedings of the {{European Conference}} on
  {{Computer Vision}} ({{ECCV}})}, 2018, pp. 164--180.

\bibitem{wangHighresolutionImageSynthesis2018}
T.-C. Wang, M.-Y. Liu, J.-Y. Zhu, A.~Tao, J.~Kautz, and B.~Catanzaro,
  ``High-resolution image synthesis and semantic manipulation with conditional
  gans,'' in \emph{Proceedings of the {{IEEE}} Conference on Computer Vision
  and Pattern Recognition}, 2018, pp. 8798--8807.

\bibitem{mechrezContextualLossImage2018}
R.~Mechrez, I.~Talmi, and L.~{Zelnik-Manor}, ``The contextual loss for image
  transformation with non-aligned data,'' in \emph{Proceedings of the
  {{European Conference}} on {{Computer Vision}} ({{ECCV}})}, 2018, pp.
  768--783.

\bibitem{benaimOnesidedUnsupervisedDomain2017}
S.~Benaim and L.~Wolf, ``One-sided unsupervised domain mapping,'' \emph{arXiv
  preprint arXiv:1706.00826}, 2017.

\bibitem{yiDualganUnsupervisedDual2017}
Z.~Yi, H.~Zhang, P.~Tan, and M.~Gong, ``Dualgan: {{Unsupervised}} dual learning
  for image-to-image translation,'' in \emph{Proceedings of the {{IEEE}}
  International Conference on Computer Vision}, 2017, pp. 2849--2857.

\bibitem{royerXganUnsupervisedImagetoimage2020}
A.~Royer, K.~Bousmalis, S.~Gouws, F.~Bertsch, I.~Mosseri, F.~Cole, and
  K.~Murphy, ``Xgan: {{Unsupervised}} image-to-image translation for
  many-to-many mappings,'' in \emph{Domain {{Adaptation}} for {{Visual
  Understanding}}}.\hskip 1em plus 0.5em minus 0.4em\relax {Springer}, 2020,
  pp. 33--49.

\bibitem{choiStarganV2Diverse2020}
Y.~Choi, Y.~Uh, J.~Yoo, and J.-W. Ha, ``Stargan v2: {{Diverse}} image synthesis
  for multiple domains,'' in \emph{Proceedings of the {{IEEE}}/{{CVF
  Conference}} on {{Computer Vision}} and {{Pattern Recognition}}}, 2020, pp.
  8188--8197.

\bibitem{liuUnsupervisedImagetoimageTranslation2017}
M.-Y. Liu, T.~Breuel, and J.~Kautz, ``Unsupervised image-to-image translation
  networks,'' in \emph{Advances in Neural Information Processing Systems},
  2017, pp. 700--708.

\bibitem{leeDiverseImagetoimageTranslation2018}
H.-Y. Lee, H.-Y. Tseng, J.-B. Huang, M.~Singh, and M.-H. Yang, ``Diverse
  image-to-image translation via disentangled representations,'' in
  \emph{Proceedings of the {{European}} Conference on Computer Vision
  ({{ECCV}})}, 2018, pp. 35--51.

\bibitem{choiStarganUnifiedGenerative2018}
Y.~Choi, M.~Choi, M.~Kim, J.-W. Ha, S.~Kim, and J.~Choo, ``Stargan: {{Unified}}
  generative adversarial networks for multi-domain image-to-image
  translation,'' in \emph{Proceedings of the {{IEEE}} Conference on Computer
  Vision and Pattern Recognition}, 2018, pp. 8789--8797.

\bibitem{jiangSaliencyGuidedImageTranslation2021}
L.~Jiang, M.~Xu, X.~Wang, and L.~Sigal,
  ``\BIBforeignlanguage{en}{Saliency-{{Guided Image Translation}}},'' p.~10,
  2021.

\bibitem{zhang2020cross}
P.~Zhang, B.~Zhang, D.~Chen, L.~Yuan, and F.~Wen, ``Cross-domain correspondence
  learning for exemplar-based image translation,'' in \emph{Proceedings of the
  IEEE/CVF Conference on Computer Vision and Pattern Recognition}, 2020, pp.
  5143--5153.

\bibitem{parkContrastiveLearningUnpaired2020}
T.~Park, A.~A. Efros, R.~Zhang, and J.-Y. Zhu, ``Contrastive learning for
  unpaired image-to-image translation,'' in \emph{European {{Conference}} on
  {{Computer Vision}}}.\hskip 1em plus 0.5em minus 0.4em\relax {Springer},
  2020, pp. 319--345.

\bibitem{johnsonPerceptualLossesRealtime2016}
J.~Johnson, A.~Alahi, and L.~{Fei-Fei}, ``Perceptual losses for real-time style
  transfer and super-resolution,'' in \emph{European Conference on Computer
  Vision}.\hskip 1em plus 0.5em minus 0.4em\relax {Springer}, 2016, pp.
  694--711.

\bibitem{parkSemanticImageSynthesis2019b}
T.~Park, M.-Y. Liu, T.-C. Wang, and J.-Y. Zhu, ``Semantic image synthesis with
  spatially-adaptive normalization,'' in \emph{Proceedings of the
  {{IEEE}}/{{CVF Conference}} on {{Computer Vision}} and {{Pattern
  Recognition}}}, 2019, pp. 2337--2346.

\bibitem{oliehoek2016concise}
F.~A. Oliehoek and C.~Amato, \emph{A concise introduction to decentralized
  POMDPs}.\hskip 1em plus 0.5em minus 0.4em\relax Springer, 2016.

\bibitem{foerster2016learning}
J.~Foerster, I.~A. Assael, N.~De~Freitas, and S.~Whiteson, ``Learning to
  communicate with deep multi-agent reinforcement learning,'' \emph{Advances in
  neural information processing systems}, vol.~29, 2016.

\bibitem{sukhbaatar2016learning}
S.~Sukhbaatar, R.~Fergus \emph{et~al.}, ``Learning multiagent communication
  with backpropagation,'' \emph{Advances in neural information processing
  systems}, vol.~29, 2016.

\bibitem{pengMultiagentBidirectionallyCoordinatedNets2017a}
P.~Peng, Y.~Wen, Y.~Yang, Q.~Yuan, Z.~Tang, H.~Long, and J.~Wang, ``Multiagent
  {{Bidirectionally-Coordinated Nets}}: {{Emergence}} of {{Human-level
  Coordination}} in {{Learning}} to {{Play StarCraft Combat Games}},'' Sep.
  2017.

\bibitem{jiang2018learning}
J.~Jiang and Z.~Lu, ``Learning attentional communication for multi-agent
  cooperation,'' \emph{Advances in neural information processing systems},
  vol.~31, 2018.

\bibitem{hoshenVAINAttentionalMultiagent2017a}
Y.~Hoshen, ``{{VAIN}}: {{Attentional Multi-agent Predictive Modeling}},'' in
  \emph{Advances in {{Neural Information Processing Systems}}}, vol.~30.\hskip
  1em plus 0.5em minus 0.4em\relax {Curran Associates, Inc.}, 2017.

\bibitem{dasTarMACTargetedMultiAgent2019a}
A.~Das, T.~Gervet, J.~Romoff, D.~Batra, D.~Parikh, M.~Rabbat, and J.~Pineau,
  ``{{TarMAC}}: {{Targeted Multi-Agent Communication}},'' in \emph{Proceedings
  of the 36th {{International Conference}} on {{Machine Learning}}}.\hskip 1em
  plus 0.5em minus 0.4em\relax {PMLR}, May 2019, pp. 1538--1546.

\bibitem{singh2018learning}
A.~Singh, T.~Jain, and S.~Sukhbaatar, ``Learning when to communicate at scale
  in multiagent cooperative and competitive tasks,'' \emph{arXiv preprint
  arXiv:1812.09755}, 2018.

\bibitem{zhengSpatiallyCorrelativeLossVarious2021}
C.~Zheng, T.-J. Cham, and J.~Cai, ``\BIBforeignlanguage{en}{The
  {{Spatially}}-{{Correlative Loss}} for {{Various Image Translation
  Tasks}}},'' \emph{\BIBforeignlanguage{en}{arXiv:2104.00854 [cs]}}, Apr. 2021.

\bibitem{deng2009imagenet}
J.~Deng, W.~Dong, R.~Socher, L.-J. Li, K.~Li, and L.~Fei-Fei, ``Imagenet: A
  large-scale hierarchical image database,'' in \emph{2009 IEEE conference on
  computer vision and pattern recognition}.\hskip 1em plus 0.5em minus
  0.4em\relax Ieee, 2009, pp. 248--255.

\bibitem{binkowski2018demystifying}
M.~Bi{\'n}kowski, D.~J. Sutherland, M.~Arbel, and A.~Gretton, ``Demystifying
  mmd gans,'' \emph{arXiv preprint arXiv:1801.01401}, 2018.

\bibitem{cazenavette2021mixergan}
G.~Cazenavette and M.~L. De~Guevara, ``Mixergan: An mlp-based architecture for
  unpaired image-to-image translation,'' \emph{arXiv preprint
  arXiv:2105.14110}, 2021.

\bibitem{kingma2014adam}
D.~P. Kingma and J.~Ba, ``Adam: A method for stochastic optimization,''
  \emph{arXiv preprint arXiv:1412.6980}, 2014.

\bibitem{he2016DeepResidual}
K.~He, X.~Zhang, S.~Ren, and J.~Sun, ``Deep residual learning for image
  recognition,'' in \emph{Proceedings of the IEEE Conference on Computer Vision
  and Pattern Recognition (CVPR)}, June 2016.

\end{thebibliography}

\newpage

 




\vfill

\end{document}